\documentclass[letterpaper, 10 pt, conference]{ieeeconf}  % Comment this line out if you need a4paper
\usepackage{times}
\usepackage{epsfig}
\usepackage{graphicx}
\usepackage{amsmath}
\usepackage{amssymb}
\usepackage{graphics} % for pdf, bitmapped graphics files
\usepackage{mathtools}
\usepackage{subcaption}
\usepackage{booktabs}
\usepackage{adjustbox}
\usepackage{caption}
\usepackage{stfloats}
\usepackage{multirow}
\usepackage[linesnumbered,ruled,lined,commentsnumbered]{algorithm2e}
\usepackage[pagebackref=true,breaklinks=true,letterpaper=true,colorlinks,bookmarks=false]{hyperref}
\usepackage{algpseudocode}
\newcommand{\dbar}{d\hspace*{-0.08em}\bar{}\hspace*{0.1em}}

\IEEEoverridecommandlockouts                              % This command is only needed if 
\title{\LARGE \bf 
Irregular Change Detection in Sparse Bi-Temporal Point Clouds using Learned Place Recognition Descriptors and Point-to-Voxel Comparison
 }

\author{Nikolaos Stathoulopoulos$^1$, Anton Koval$^1$ and George Nikolakopoulos$^1$ % <-this % stops a space
\thanks{This work has been funded by the European Unions Horizon 2020 Research and Innovation Programme under the Grant Agreement No. 101003591 NEX-GEN SIMS.}% <-this % stops a space
\thanks{$^{1}$The Authors are with the Robotics and AI Group, Department of Computer, Electrical and Space Engineering, Lule\r{a} University of Technology, 971 87 Lule\r{a}, Sweden} %
\thanks{Corresponding Author's Email: \texttt{niksta@ltu.se}} %
}

\begin{document}

\maketitle
\thispagestyle{empty}
\pagestyle{empty}
%%%%%%%%%%%%%%%%%%%%%%%%%%%%%%%%%%%%%%%%%%%%%%%%%%%%%%%%%%%%%%%%%%%%%%%%%%%%%%%%
%%%%%%%%%%%%%%%%%%%%%%%%%%%%%%%%%%%% ABSTRACT %%%%%%%%%%%%%%%%%%%%%%%%%%%%%%%%%%
\begin{abstract}
   Change detection and irregular object extraction in 3D point clouds is a challenging task that is of high importance not only for autonomous navigation but also for updating existing digital twin models of various industrial environments. This article proposes an innovative approach for change detection in 3D point clouds using deep learned place recognition descriptors and irregular object extraction based on voxel-to-point comparison. The proposed method first aligns the bi-temporal point clouds using a map-merging algorithm in order to establish a common coordinate frame. Then, it utilizes deep learning techniques to extract robust and discriminative features from the 3D point cloud scans, which are used to detect changes between consecutive point cloud frames and therefore find the changed areas. Finally, the altered areas are sampled and compared between the two time instances to extract any obstructions that caused the area to change. The proposed method was successfully evaluated in real-world field experiments, where it was able to detect different types of changes in 3D point clouds, such as object or muck-pile addition and displacement, showcasing the effectiveness of the approach. The results of this study demonstrate important implications for various applications, including safety and security monitoring in construction sites, mapping and exploration and suggests potential future research directions in this field.
\end{abstract}

%%%%%%%%%%%%%%%%%%%%%%%%%%%%%%%%%%%%%%%%%%%%%%%%%%%%%%%%%%%%%%%%%%%%%%%%%%%%%%%%
%%%%%%%%%%%%%%%%%%%%%%%% INTRODUCTION %%%%%%%%%%%%%%%%%%%%%%%%%%%%%%%%
\section{Introduction} \label{sec:introduction}

Three-dimensional point cloud change detection is a powerful tool for identifying changes in a 3D environment, and is often used to track changes over time~\cite{approaches}. The process typically involves comparing two or more point clouds, and identifying differences between them. This can be done using various methods, such as feature-based registration~\cite{feature_based}, point-based registration~\cite{point_based} and semantic segmentation~\cite{DBLP:journals/corr/abs-2108-06103}. Feature-based registration involves identifying and matching features in the point clouds, such as corners or edges, to align the point clouds and identify changes. Point-based registration involves aligning the point clouds based on the individual points, without the need for identifying specific features. Semantic segmentation~\cite{semantic} involves first clustering and then classifying the points in the point cloud into different categories, such as buildings, roads, and trees, to identify changes in the environment. 
Change detection in 3D point clouds is a challenging task that has gained significant attention in the past few years due to its various applications in areas, such as autonomous navigation~\cite{KHATAB202136}, robotics, surveying and construction~\cite{rs10101512}, industrial inspection~\cite{inpsection}, environmental monitoring~\cite{trees}, urban planning~\cite{DUEKER197289}, security and surveillance and so on. For example, in autonomous vehicles, point cloud change detection can be utilized to track changes in the environment, such as new obstacles or road closures~\cite{Ku2021}. For more specific applications in robotics that involve grasping or manipulation, change detection can be used to track the movement of objects~\cite{photonics8090394}, allowing the robot to adapt its actions accordingly. In surveying and construction, 3D point cloud change detection can be used to track changes in a construction site over time, allowing for accurate monitoring of progress and identification of any potential issues. Similarly, it can be used in industrial inspection to detect changes in the condition of equipment, structures, and infrastructure, such as cracks, corrosion and deformations. In environmental monitoring, it can be used to track changes in vegetation, land use, and natural disasters such as floods, landslides, and wildfires while in security and surveillance applications it can be used to detect changes in a scene, such as the presence of new objects or people, and alert security personnel to potential threats. Overall, 3D point cloud change detection is a versatile and valuable tool that can be applied to a wide range of fields to track changes in 3D environments, allowing for more accurate and efficient decision-making. 
%%%%%%%%%%%%%%%%%%%%%%%%%%%%%%%%%%%%%%%%%%%%%%%%%%%%%%%%%%%%%%%%%%%%%%%%%%%%%%%%
\begin{figure}[t!] 
    \includegraphics[width=1.\columnwidth]{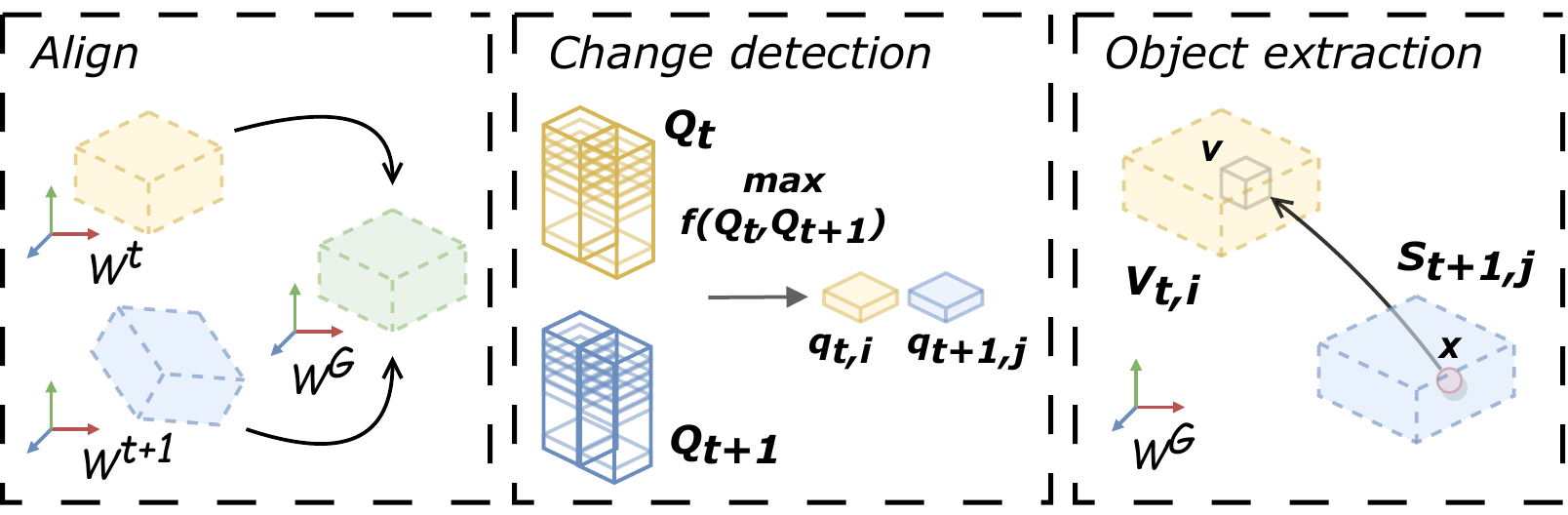}
    \caption{The presented framework consists of three main steps. Aligning the bi-temporal maps into a common global frame $\mathcal{W}^G$, finding the changed areas $S_{t,i}, S_{t+1,j}$ by extracting the regions with the highest distance in the descriptors' $Q_t$ and $Q_{t+1}$ vector space and finally extracting the objects $O_n$ by comparing if the points $x$ of one area are included in the corresponding voxels $v$ in the other voxelized area.} \label{fig:concept_figure}
    \vspace{-7.mm}
\end{figure} 
%%%%%%%%%%%%%%%%%%%%%%%%%%%%%%%%%%%%%%%%%%%%%%%%%%%%%%%%%%%%%%%%%%%%%%%%%%%%%%%%

The traditional approaches for change detection rely on feature-based methods and registration techniques. However, these methods are often sensitive and are heavily affected by noise, outliers, different point densities and distributions~\cite{AFAQ2021101310}, making it difficult to achieve accurate and robust change detection. One way to improve the accuracy of such methods is to apply filtering and denoising to the point cloud data. On the other hand, in recent years, there has been a growing interest in using 3D point cloud change detection in combination with machine learning and deep learning techniques~\cite{DBLP:journals/corr/abs-2006-05612}, which can provide more robust and accurate results in comparison with traditional methods. In this context, deep learning techniques have emerged as a powerful tool for change detection in 3D point clouds. These techniques are able to extract robust and discriminative features from 3D point clouds, which can then be used to detect changes between consecutive point cloud frames. In this article, we propose a novel approach, briefly described on Fig.~\ref{fig:concept_figure}, for change detection in sparse 3D point clouds using deep learned place recognition descriptors and object extraction based on point-to-voxel comparison and evaluate it through real-world field experiments, in harsh environments. % Explain our framework a bit more, highlighting in which part of change detection we are contributing
%%%%%%%%%%%%%%%%%%%%%%%%%%%%%%%%%%%%%%%%%%%%%%%%%%%%%%%%%%%%%%%%%%%%%%%%%%%%%%%%

%%%%%%%%%%%%%%%%%%%%%%%%%%%%%% CONTRIBUTIONS %%%%%%%%%%%%%%%%%%%%%%%%%%%%%%%%%%%
\subsection*{\textit{— Contributions}}
The contributions of the proposed framework can be summarized as follows: (a) An approach that utilizes deep learned place recognition descriptors to detect changes in sparse 3D point clouds by reverting the problem of searching for the minimum distance between descriptors, which is a novel approach that is absent from the existing literature. (b) The proposed framework employs point-to-voxel comparison on the bi-temporal point clouds, after ensuring a common coordinate frame, which allows for a fast and efficient abnormality extraction from the changed areas, in contrast to other methods that rely on edge or planar extraction~\cite{semantic, semantic_2}. (c) The framework is designed to be scalable, as it offers fast processing time and can be used for large-scale environments. This allows the framework to efficiently narrow down the changed areas and extract changes. (d) The learned approach used in the proposed framework enables the detection of changed areas in real-time, which is an improvement over the time-consuming geometrical approaches~\cite{feature_based, AFAQ2021101310}. (e)  The field evaluation results of the proposed framework show promising results for real-world applicability in inspection and monitoring scenarios. This highlights the potential of the proposed framework for use in various real-world applications. 

Overall, these contributions demonstrate the significance of the proposed framework for 3D point cloud change detection and irregular object extraction and highlight its potential for field applications. The rest of the article is structured as it follows. Section~\ref{sec:related_work} provides an overview of related works in the field, offering insight into the context of the study. Section~\ref{sec:formulation} delves into the problem formulation, clearly defining the problem being addressed and setting the stage for the methodology to be presented in Section~\ref{sec:methodology}. The experimental evaluation, including the results of the study, can be found in Section~\ref{sec:experimental_evaluation}. Finally, Section~\ref{sec:conclusions} concludes the article by summarizing the main findings and contributions of the study, as well as discussing directions for future work. 

%%%%%%%%%%%%%%%%%%%%%%%%%%%%%%%%%%%%%%%%%%%%%%%%%%%%%%%%%%%%%%%%%%%%%%%%%%%%%%%%
%%%%%%%%%%%%%%%%%%%%%%%%%%%%%%% RELATED WORK %%%6%%%%%%%%%%%%%%%%%%%%%%%%%%%%%%%%
\section{Related work} \label{sec:related_work}

%%%%%%%%%%%%%%%%%%%%%%%%%%%%%%%%%%%%%%%%%%%%%%%%%%%%%%%%%%%%%%%%%%%%%%%%%%%%%%%%
\begin{figure*}[t!]
    \includegraphics[width=1.\textwidth]{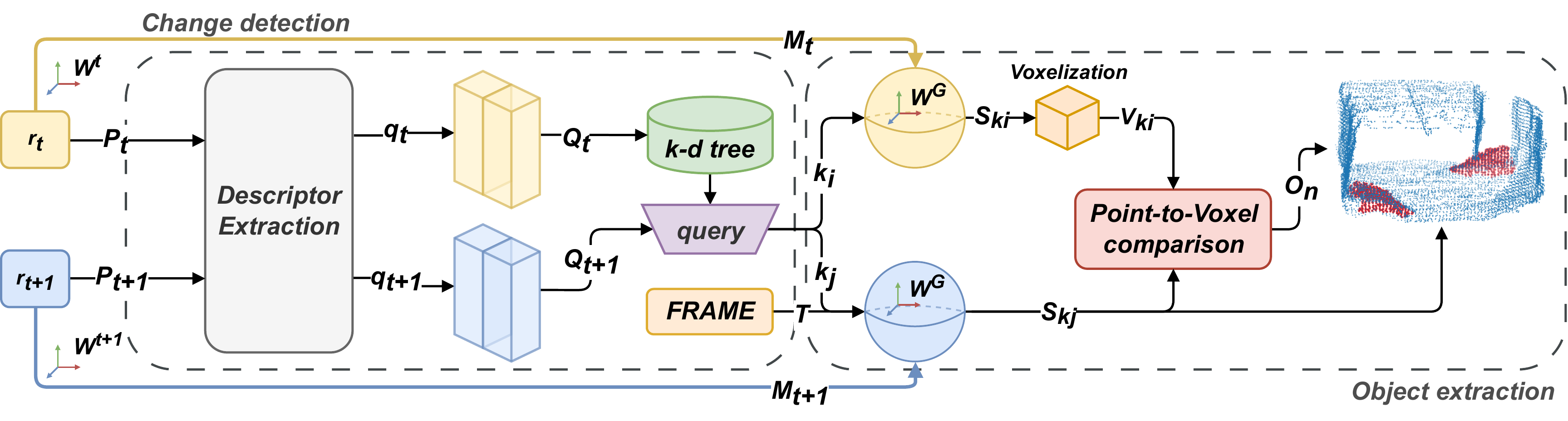}
    \caption{The overall change detection and object extraction pipeline of the proposed solution. The robot $r$ explores or navigates through the same area in two different time instances, $t$ and $t+1$, while simultaneously collecting descriptors in a stack $Q$ from the point clouds $P$. After aligning both maps $M_t$ and $M_{t+1}$ to the common map frame $\mathcal{W}^G$, the changed areas $S_{k_i}$ and $S_{k_j}$ are detected using the place recognition descriptors $Q_t$ and $Q_{t+1}$ and then within those areas the objects are extracted by a point-to-voxel comparison.}
    \vspace{-6mm}
    \label{fig:architecture}
\end{figure*}
%%%%%%%%%%%%%%%%%%%%%%%%%%%%%%%%%%%%%%%%%%%%%%%%%%%%%%%%%%%%%%%%%%%%%%%%%%%%%%%%

In the present literature, the problem of change detection is commonly addressed using multi-temporal images~\cite{electronics11030431, msgatn} and more specifically using object-based methods due to their high performance and robustness. However, these methods often struggle to capture discriminative features, particularly when dealing with irregular objects. To address this challenge, Shuai et al.~\cite{msgatn} proposed a novel method called the superpixel-based multiscale Siamese graph attention network (MSGATN), which addresses the difficulty of capturing discriminative features in irregular objects by using superpixel segmentation to divide images into homogeneous regions. The method then models the problem using graph theory to construct a series of nodes with the adjacency between them and extracts multiscale neighborhood features through applying a graph convolutional network and an attention mechanism. The final binary change map is obtained by classifying each node using fully connected layers. The proposed method is shown to be effective and efficient in detecting changes in real datasets and can greatly reduce the dependence on labeled training samples in a semi-supervised training fashion.

Apart from multi-temporal images, the change detection problem can be addressed with methods based on the consistency between the occupancy of space, computed from different datasets. In~\cite{cd_mls}, a new approach for detecting changes in urban areas using mobile laser scanning (MLS) point clouds is presented. By using the Weighted Dempster-Shafer theory (WDST) to fuse the occupancy of scan rays and compare the results with a conventional point to triangle (PTT) distance method, the proposed approach is fully automatic and allows for the detection of changes at large scales in urban scenes with fine detail, and the ability to distinguish real changes from occlusions. Other recent work~\cite{Gojcic2021} introduces solutions that estimate 3D displacement vectors directly from point cloud data. The solution is fully automated and through the main pipeline, it searches for corresponding points across different epochs in the space of 3D local feature descriptors. This approach is different from traditional methods that establish displacements based on proximity in Euclidean space and is able to detect motion and deformations that occur parallel to the underlying surface. The proposed solution proves to be efficient and can be applied to point clouds of any size. 
Last but not least, there has been a growing interest in using machine learning and deep learning techniques for 3D point cloud change detection. These methods have been shown to be more robust and accurate than traditional methods, and can be used in a wide range of applications. An example of this type of work can be found in~\cite{dpdist}. The authors propose a new deep learning method for point cloud comparison called Deep Point Cloud Distance (DPDist). The approach measures the distance between points in one cloud and the estimated surface from which the other point cloud is sampled. The surface is estimated locally and efficiently using a 3D modified Fisher vector representation. This local representation reduces the complexity of the surface, allowing for efficient and effective learning that generalizes well across object categories. The method is tested on challenging tasks such as similar object comparison and registration and it is shown to provide significant improvements over commonly used distances such as Chamfer distance~\cite{chamfer}, Earth mover's distance~\cite{earth}, and others.

Inspired by the applicability of machine learning techniques in the field of autonomous robots, we propose a novel framework, that utilizes place recognition techniques, such as global descriptors, to detect changes between two bi-temporal point cloud maps. This approach is particularly useful in  dynamic scenes and large-scale settings where changed areas can be narrowed down. 

%%%%%%%%%%%%%%%%%%%%%%%%%%%%%%%%%%%%%%%%%%%%%%%%%%%%%%%%%%%%%%%%%%%%%%%%%%%%%%%%
\section{Problem formulation}\label{sec:formulation}

The main focus of this research is to develop a framework able to find one or more changed regions and extract the potential objects that caused the abnormality, given two point cloud maps from different time periods. By narrowing down the search area to only the changed regions, any objects that caused the change can be extracted faster, since there is no need to compare the entire point cloud maps. Considering a robot $r$ operating in $\mathbb{R}^3$ space, in an iteration $t$, it generates a point cloud map $M_t$, with respect to its local static coordinate frame $\mathcal{W}^t$. The map can be defined as a set of points $m \in \mathbb{R}^3$:
\begin{equation}
    M_t = \{m_{1}, m_{2}, \ldots, m_{N_t}\} \:\: \text{with} \:\: N_t \in \mathbb{N}
\end{equation}
Similarly, the trajectory of the robot can be defined as a set of points $p \in \mathbb{R}^3$:
\begin{equation}
    Tr_t = \{p_k | k = 1,2,\ldots,K_t\} \:\: \text{with} \:\: K_t \in \mathbb{N},
\end{equation}
where $p_k = (x_k, y_k, z_k)_t$ is the position of the robot $r$ at a time step $k$ relative to its local frame $\mathcal{W}^t$. Therefore, given two maps from two different mapping iterations, $M_t$ and $M_{t+1}$, so that $M_{t} \cup M_{t+1} \neq \O$, we need to determine the pair of regions that have the most change between them and hence we define the problem as:
\begin{equation} \label{eq:change}
    \operatorname*{max}_{(i,j) \, \in \, \mathbb{N}} \dbar(S_{k_i}, S_{k_j}),
\end{equation}
where $S$ is a subset sphere of the map $M$ and is denoted as:
\begin{equation}
    S = \{m,p \in \mathbb{R}^3: ||m-p_k||^2 \leq r^2\} \subset M
\end{equation}
To acquire the spheres $S$ we sample the point cloud map $M$ from a trajectory point $p_k \in Tr$. The function $\dbar(X, Y)$ represents the difference between two regions $X$ and $Y$. Furthermore, after acquiring the regions $S_k$, the objects of interest $O_n \in \mathbb{R}^3$ should be extracted, which based on the set theory can be denoted as: 
\begin{equation}
    O_n = S_{k_j} \symbol{92} S_{k_i} = \{x \in S_{k_j} : x \not\in S_{k_i}\}
\end{equation}
Respectively the object can be part of $S_{k_i}$ instead of $S_{k_j}$. As an object of interest, we define any object or obstruction that was either not present in the first point cloud map or has been displaced, thus causing the area to change.
%%%%%%%%%%%%%%%%%%%%%%%%%%%%%%%%%%%%%%%%%%%%%%%%%%%%%%%%%%%%%%%%%%%%%%%%%%%%%%%%
%%%%%%%%%%%%%%%%%%%%%%%%%%%%%%%%%%%%%%%%%%%%%%%%%%%%%%%%%%%%%%%%%%%%%%%%%%%%%%%%
\begin{figure}[b!]
    \centering
    \includegraphics[width=0.75\columnwidth]{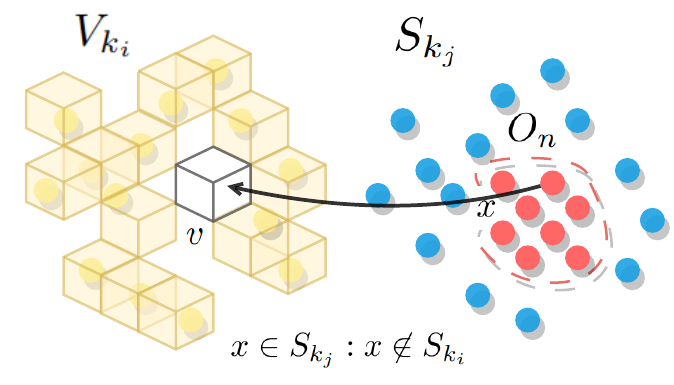}
    \caption{The object extraction process to satisfy (Eq.~\ref{eq:change}) in a fast and efficient way is to check every point $x \in S_{k_j}$ if it is included in any of the voxels $v \in V_{k_i}$.}
    \label{fig:object_extraction}
\end{figure}
%%%%%%%%%%%%%%%%%%%%%%%%%%%%%%%%%%%%%%%%%%%%%%%%%%%%%%%%%%%%%%%%%%%%%%%%%%%%%%%%
% \vspace{-0.25cm}
%%%%%%%%%%%%%%%%%%%%%%%%%%%%%%%%%%%%%%%%%%%%%%%%%%%%%%%%%%%%%%%%%%%%%%%%%%%%%%%%
\begin{figure*}[b!]
  \begin{subfigure}{0.9\columnwidth}
    \includegraphics[width=\columnwidth]{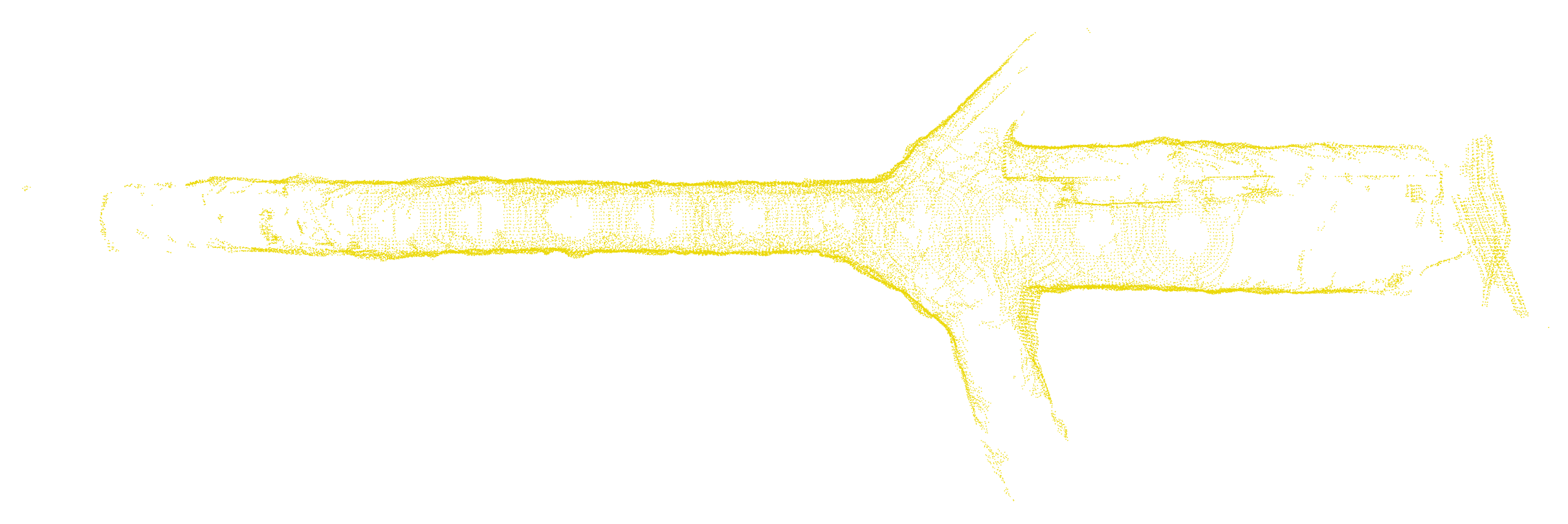}
    \caption{Map $M_t$}
    \end{subfigure}
    \hfill
    \begin{subfigure}{0.9\columnwidth} 
    \includegraphics[width=\columnwidth]{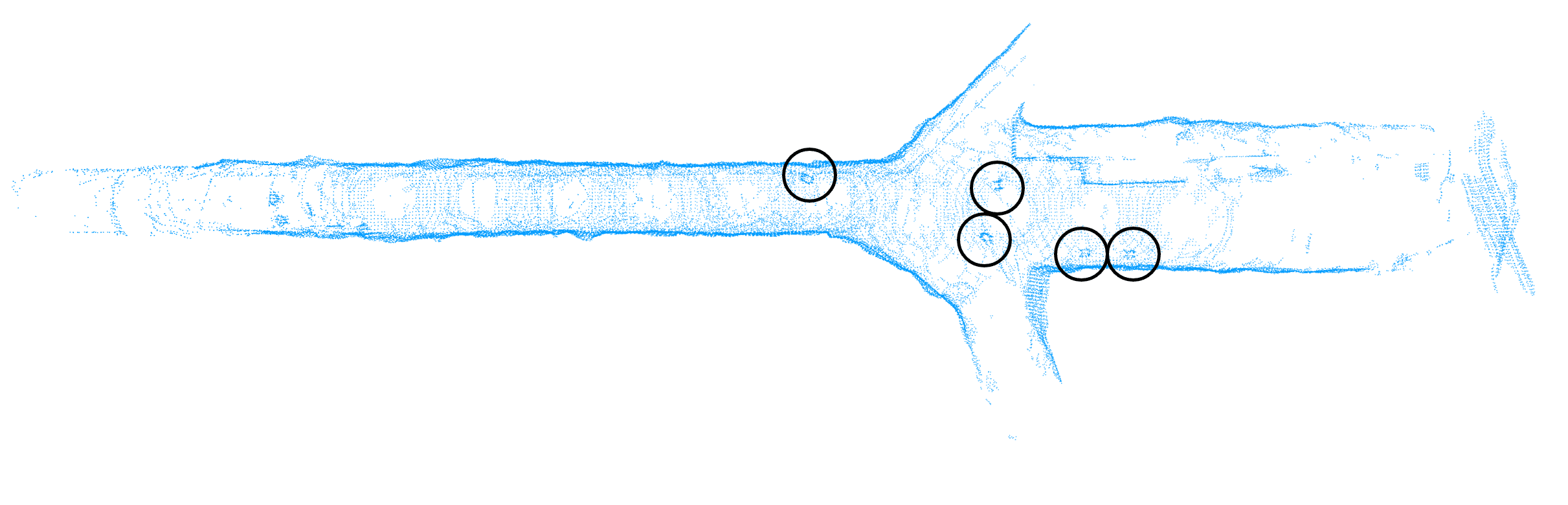}
    \caption{Map $M_{t+1}$}
    \end{subfigure}
    % \vspace{5pt} \\
    \begin{subfigure}{0.9\columnwidth} 
    \includegraphics[width=\columnwidth]{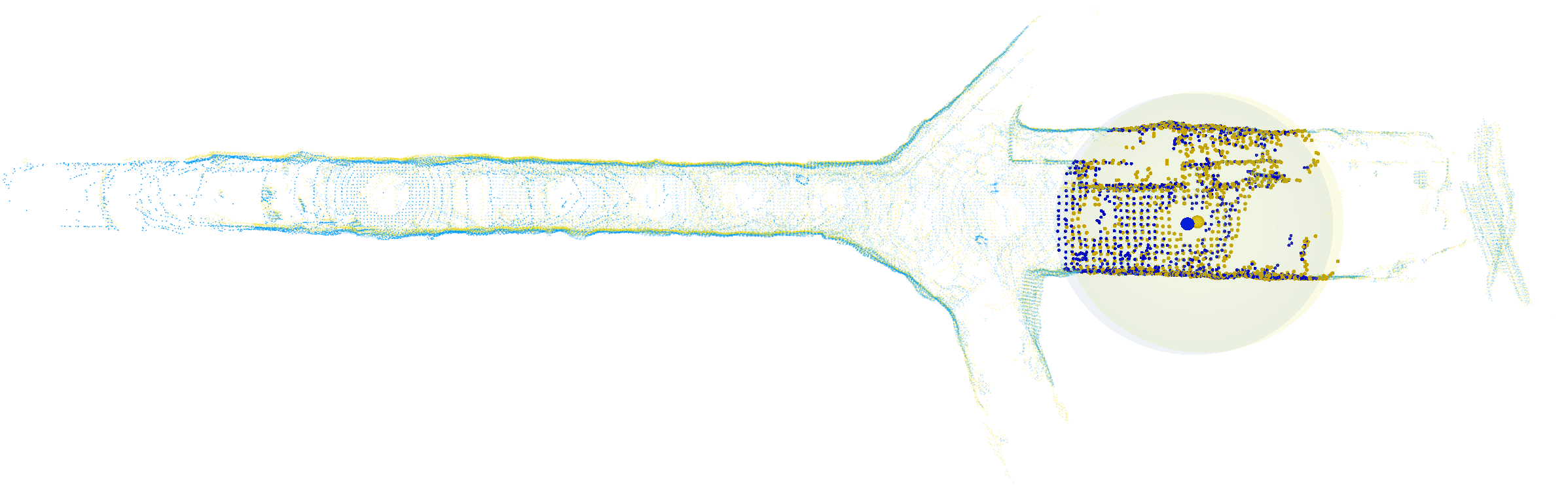} 
    \caption{After alignment} \label{subfig:changes_mjolk}
    \end{subfigure}
    \hfill
    \begin{subfigure}{0.9\columnwidth}
    \includegraphics[width=\columnwidth]{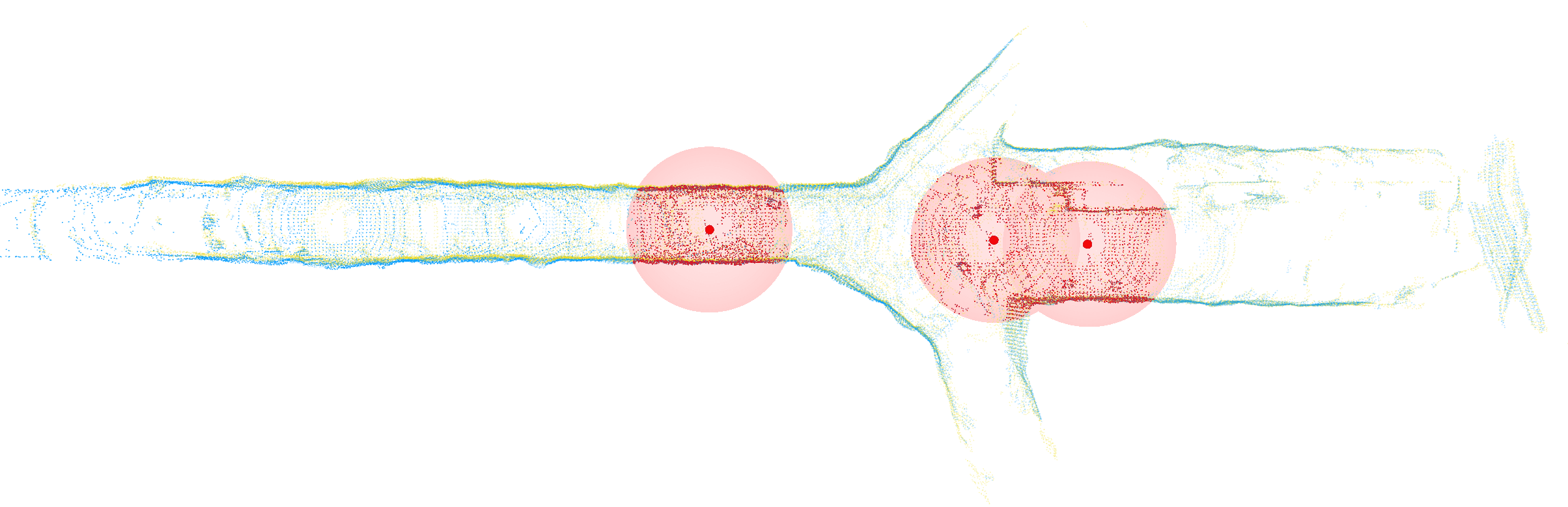}
    \caption{Detected changed areas} \label{subfig:aligned_mjolk}
    \end{subfigure}
    \caption{Subfigure (a) depicts the map $M_t$ before the changes, consisting of a total of $53,117$ points, while subfigure (b) depicts the map $M_{t+1}$ after the changes, highlighted with a black circle, consisting of $63,942$ points. Subfigure (c) depicts the two maps after the alignment and subfigure (d) demonstrates the detected changed areas in red color.}
    \label{fig:maps_mjolkberget}
    % \vspace{-12mm}
\end{figure*}
%%%%%%%%%%%%%%%%%%%%%%%%%%%%%%%%%%%%%%%%%%%%%%%%%%%%%%%%%%%%%%%%%%%%%%%%%%%%%%%%

\section{Methodology} \label{sec:methodology}

An overview of the proposed 3D point cloud change detection framework is given briefly in Fig.~\ref{fig:architecture}. The first step to tackle the aforementioned problem (\ref{sec:formulation}) is to align the two maps $M_t$ and $M_{t+1}$, since the starting position of the robot is never exactly the same and therefore the global coordinate frames $\mathcal{W}^t$ and $\mathcal{W}^{t+1}$ will have an offset $T$. To find this offset, the first module of the pipeline is the map merging framework FRAME~\cite{Stathoulopoulos2023frame}. By utilizing the pairs of maps and trajectories, $M_t, M_{t+1}$ and $Tr_t, Tr_{t+1}$, 
% as well as the place recognition, deep learned descriptors 3DEG~\cite{3DEG}, 
FRAME is able to provide a homogeneous rigid transformation of the special Euclidean group $T:\mathbb{R}^3 \rightarrow \mathbb{R}^3$, denoted as:
\begin{equation} \label{eq:transform}
    T = \left[ \begin{array}{cc}
         R & p\\
         0 & 1
    \end{array} \right] \in SE(3),
\end{equation}
where $R \in SO(3)$ and $p \in \mathbb{R}^3$. The map merging process can be described as a function $f_m: \mathbb{R}^3 \times \mathbb{R}^3 \rightarrow \mathbb{R}^3$, or more precisely:
\begin{equation} 
    M = f_m(M_t,M_{t+1}) = M_t \cup T \, M_{t+1}
\end{equation}
After the first step, both maps are with respect to the same coordinate frame $\mathcal{W}^G$, making the handling of the points straightforward. 
The subsequent step is to find the changed regions. Based on Eq.~\eqref{eq:change}, finding the indexes $i$ and $j$ from the trajectories $Tr_t$ and $Tr_{t+1}$, would yield the center points to sample the spherical regions $S_{k_i}$ and $S_{k_j}$. To do so, we utilize the same place recognition, deep learned descriptors used by the map merging framework. The 3DEG~\cite{3DEG} descriptors are a set of place recognition and yaw discrepancy regression vectors, $\Vec{q}$ and $\Vec{w}$ respectively. More specifically, vector $\Vec{q}$ is an orientation-invariant, place dependent vector used for querying similar point clouds and $\Vec{w}$ is an orientation-specific vector, used for reducing the yaw difference between two point cloud scans. The vectors are acquired and saved as the robot maps the area and are ready to be used in the next iteration of mapping. We define the saved vector sets as:
\begin{gather}
    Q = \{\vec q \in \mathbb{R}^{64}, \, K_t \in \mathbb{N}: \vec q_1, \vec q_2, \dots, \vec q_{K_t} \} \\
    W = \{\vec w \in \mathbb{R}^{64}, \, K_t \in \mathbb{N}: \vec w_1, \vec w_2, \dots, \vec w_{K_t} \}
\end{gather}
In this case, since the two maps are already aligned, we will only make use of the place dependent vector set $Q$. To find the regions with the most change, we invert the process of querying the vector set to get the most similar point clouds. A \textit{k-d} tree is built using the first vector set $Q_t$ and then queried with the other vector set $Q_{t+1}$. Consequently, we can find the pair of vectors $\vec q_{t,i}$ and $\vec q_{t+1,j}$ that have the maximum distance between them in the new vector space and therefore contain the changes. The aforementioned can be described as:
\begin{equation}
    (k_i,k_j) = \operatorname*{arg\,max}_{(i,j) \, \in \, \mathbb{N}} f (Q_{t,i},Q_{t+1,j})
\end{equation}
After having both spheres $S_{k_i}$ and $S_{k_j}$ the next step would be to extract the irregular object $O_n$. Considering the points $m_{k_i} \in S_{k_i}$ and $m_{k_j} \in S_{k_j}$, we can extract the object of interest by comparing the amount of neighbors $N$ of a point in the changed region, $x \in S_{k_j}$, with respect to the unchanged region $S_{k_i}$, since they both have the same coordinate frame. In order to avoid the $O(n^2)$ complexity of this approach, we opt for voxelizing the sphere $S_{k_i}$ and checking if the points of the sphere $S_{k_j}$ are included in the voxels $v$ of the voxelized sphere $V_{k_i}$. This is denoted as:
\begin{equation}
    O_n = \{x \in S_{kj} \; | \; \forall \ v \in V_{k_i}, \ x \not\in v\},
\end{equation}
and depicted on Fig.~\ref{fig:object_extraction}. The voxelized sphere $V_{k_i}$ represents an occupancy grid, where a voxel $v$ is set to occupied if there are enough points populating it. As a final step, a statistical outlier removal filter is applied to the object $O_n$ to remove any noise or misclassified points. Formally, this process can be expressed as follows. Let $X = \{x_1, x_2, ..., x_n\}$ be a set of data points, with a mean of $\mu$ and a standard deviation of $\sigma$. Outliers are considered the points that fall outside of the interval $\left(\mu - \lambda\sigma, \mu + \lambda\sigma\right)$, for a preset threshold $\lambda$.
% \begin{equation}
%     O_n = \biggl\{ x \in \mathbb{R}^3 : \sum_{x \in S_{k_j}} \sum_{v \in V\left[S_{k_i}\right]} \Bigl[ g(x,m) \Bigr] \leq N_{th} \biggr\},
% \end{equation}
% where $N_{th}$ is the threshold of neighbors to consider a point part of the object. In theory $N_{th}$ should be zero, but to accommodate for any noise in the point cloud scan, we set a small threshold. The Iverson bracket defines the condition for a point to be considered a neighbor.
% \begin{equation}
%     \Bigl[ g(x,m) \Bigr] = \left\{
%         \begin{array}{ll}
%         1, \; \text{if} \; \| x -  m\|^2  \leq r_n^2 \\
%         0, \; \text{otherwise}
%         \end{array} 
%     \right.,
% \end{equation}
% where $r_n$ is the radius of the neighborhood. 

%%%%%%%%%%%%%%%%%%%%%%%%%%%%%%%%%%%%%%%%%%%%%%%%%%%%%%%%%%%%%%%%%%%%%%%%%%%%%%%%

\section{Experimental Evaluation} \label{sec:experimental_evaluation}

The experimental evaluation section of this article provides a comprehensive analysis of the performance of the proposed change detection and object extraction method. The goal of this section is to demonstrate the effectiveness and robustness of the method in a real-world scenario, while providing a detailed discussion of the results, including quantitative and qualitative analyses of the performance, in order to support the conclusions and provide insights into the strengths and limitations of the proposed approach.

%%%%%%%%%%%%%%%%%%%%%%%%%%%%%%%%%%%%%%%%%%%%%%%%%%%%%%%%%%%%%%%%%%%%%%%%%%%%%%%%
\begin{figure*}[b!]
  \begin{subfigure}{0.28\textwidth}
    \includegraphics[width=\textwidth]{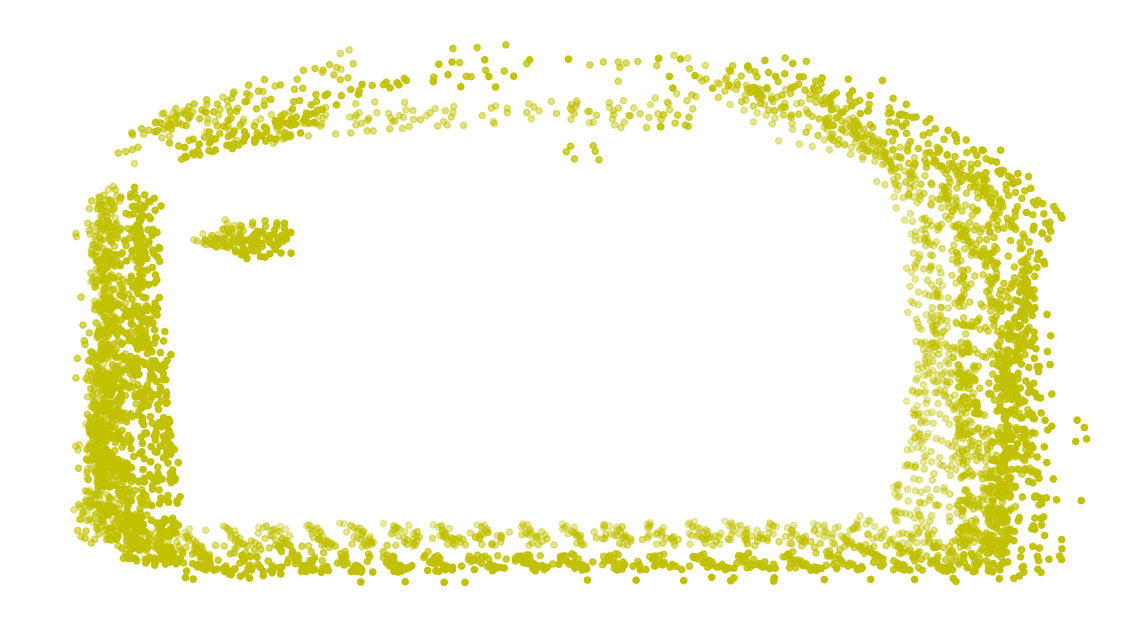}
    \caption{} \label{subfig:unchanged_1_mjolk}
    \end{subfigure}
    \hfill
    \begin{subfigure}{0.28\textwidth}
    \includegraphics[width=\textwidth]{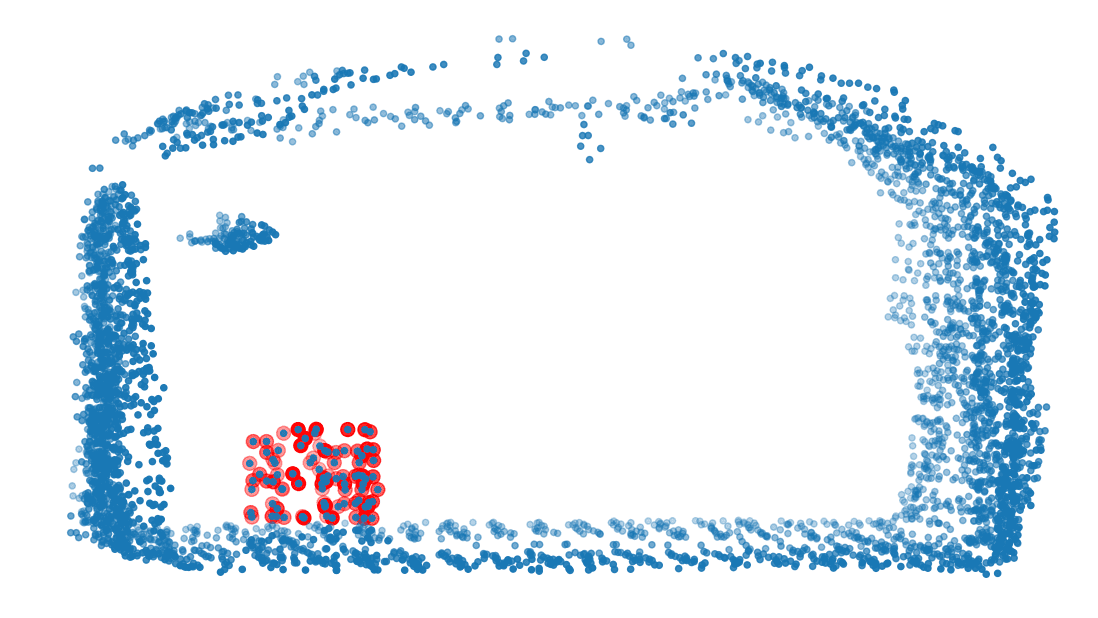}
     \caption{} \label{subfig:changed_1_mjolk}
    \end{subfigure}
   \begin{subfigure}{0.21\textwidth}
    \includegraphics[width=\textwidth]{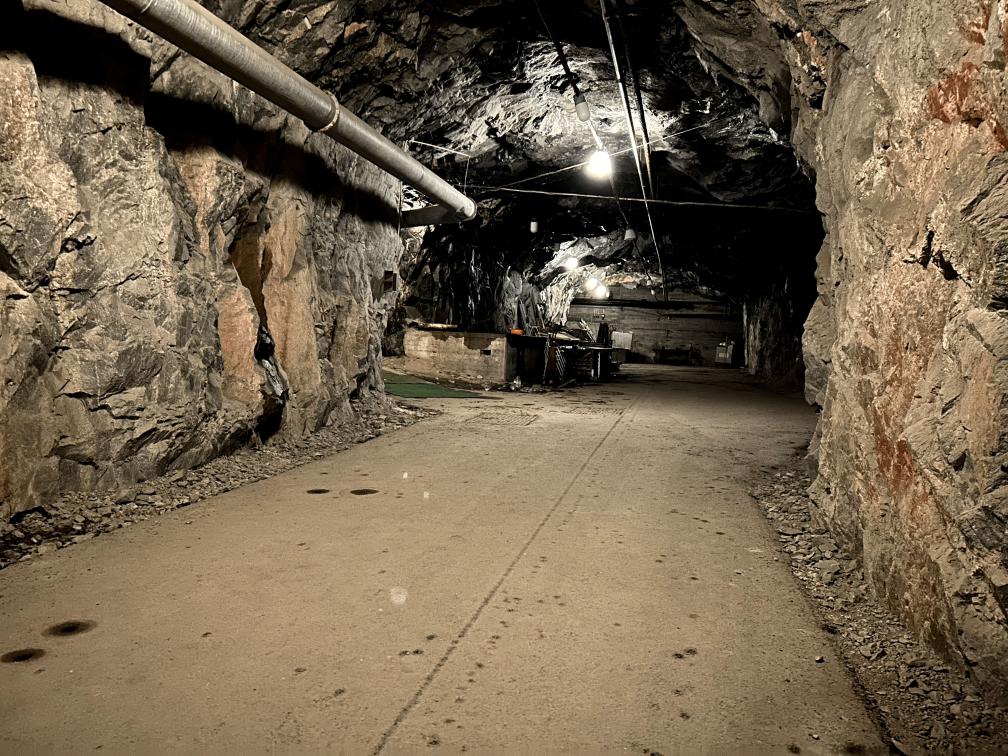}
    \caption{} 
    \end{subfigure}
    \begin{subfigure}{0.21\textwidth}
    \includegraphics[width=\textwidth]{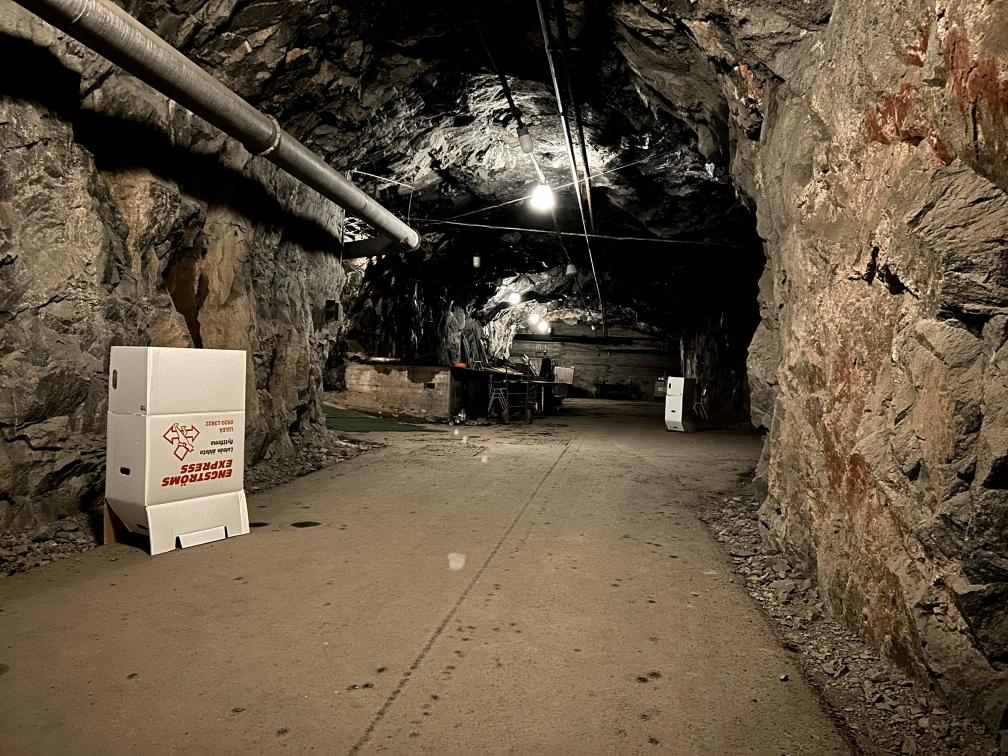}
     \caption{}
    \end{subfigure}
    \vspace{5pt} \\
    \begin{subfigure}{0.28\textwidth}
    \includegraphics[width=\textwidth]{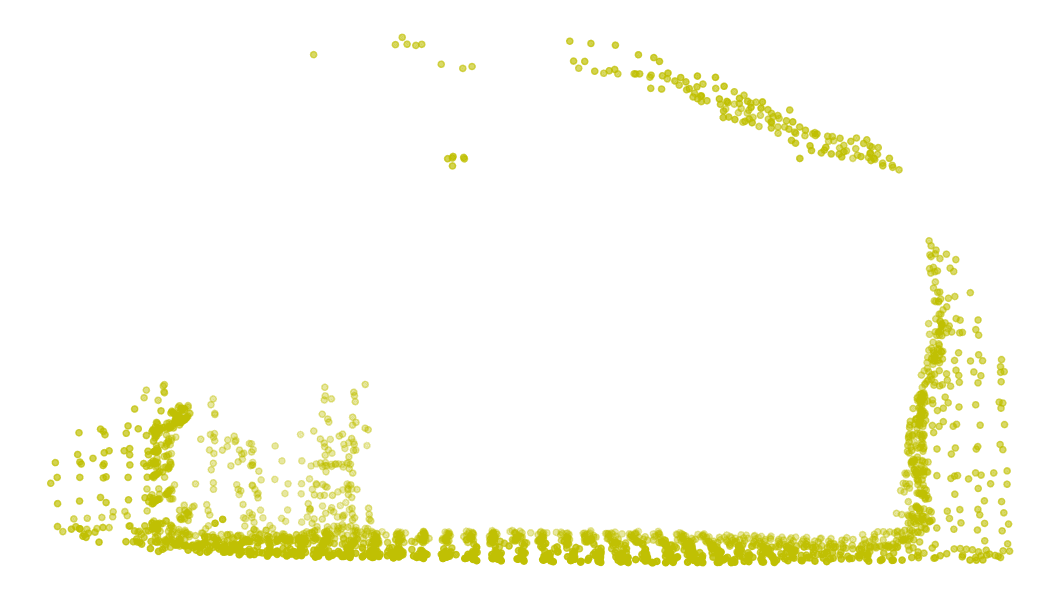} 
    \caption{}  \label{subfig:unchanged_2_mjolk}
    \end{subfigure}
    \begin{subfigure}{0.28\textwidth}
    \includegraphics[width=\textwidth]{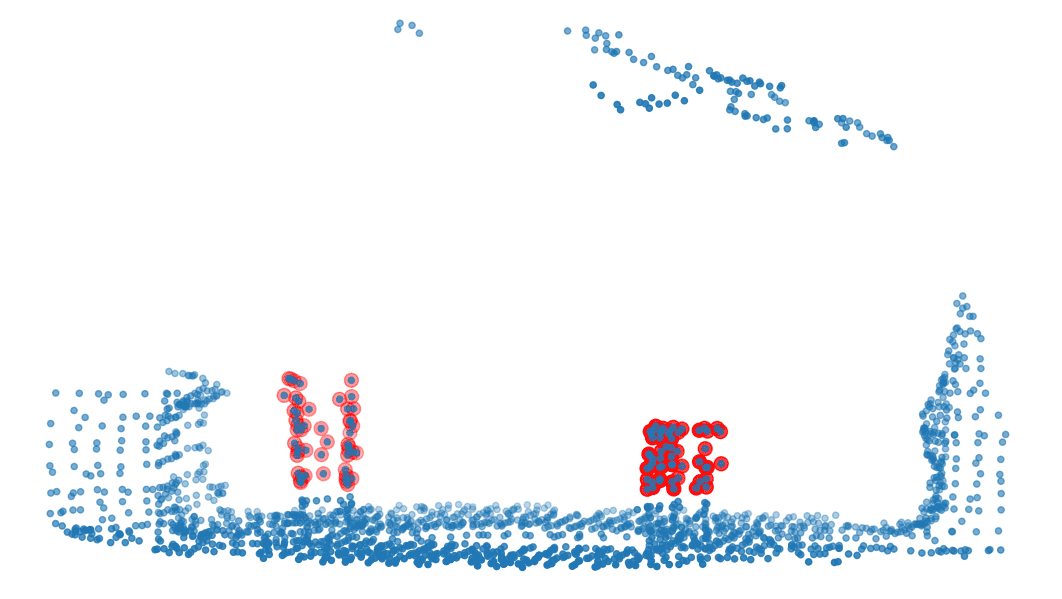} 
    \caption{} \label{subfig:changed_2_mjolk}
    \end{subfigure}
    \begin{subfigure}{0.21\textwidth}
    \includegraphics[width=\textwidth]{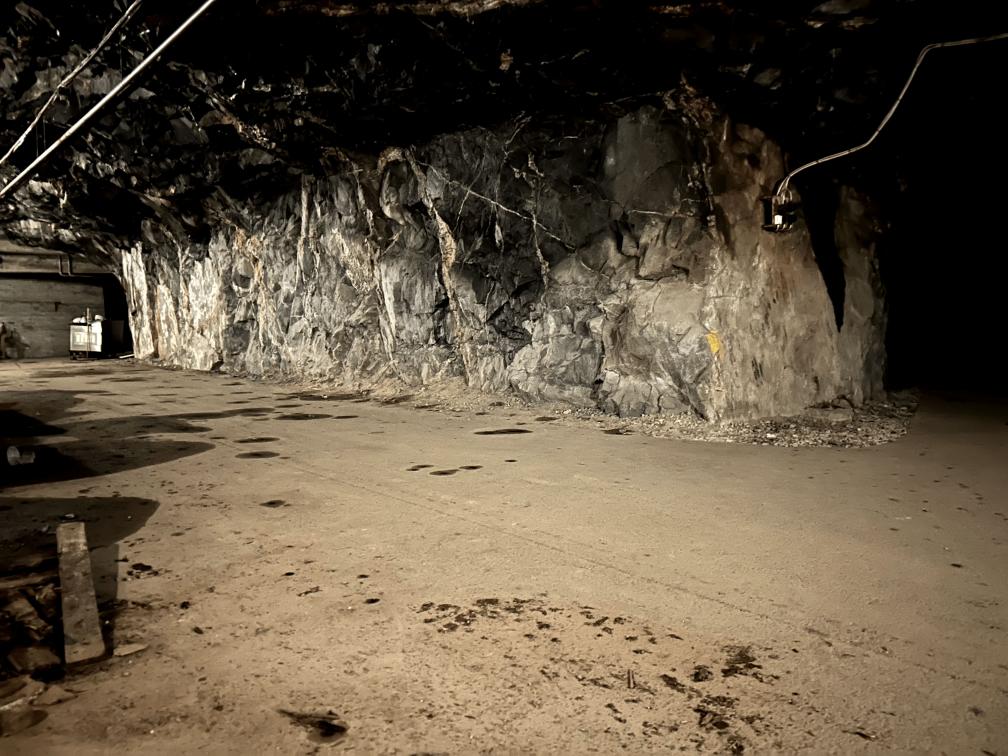}
    \caption{}
    \end{subfigure}
    \hfill
    \begin{subfigure}{0.21\textwidth}
    \includegraphics[width=\textwidth]{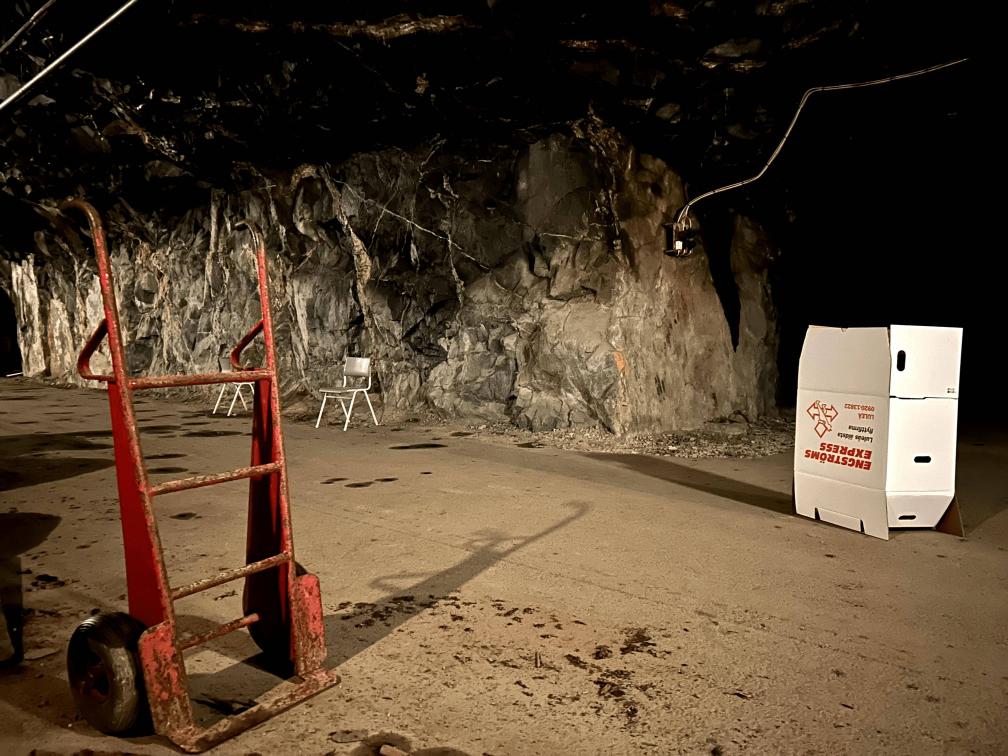}
     \caption{} 
    \end{subfigure}
    \vspace{5pt} \\
    \begin{subfigure}{0.28\textwidth}
    \includegraphics[width=\textwidth]{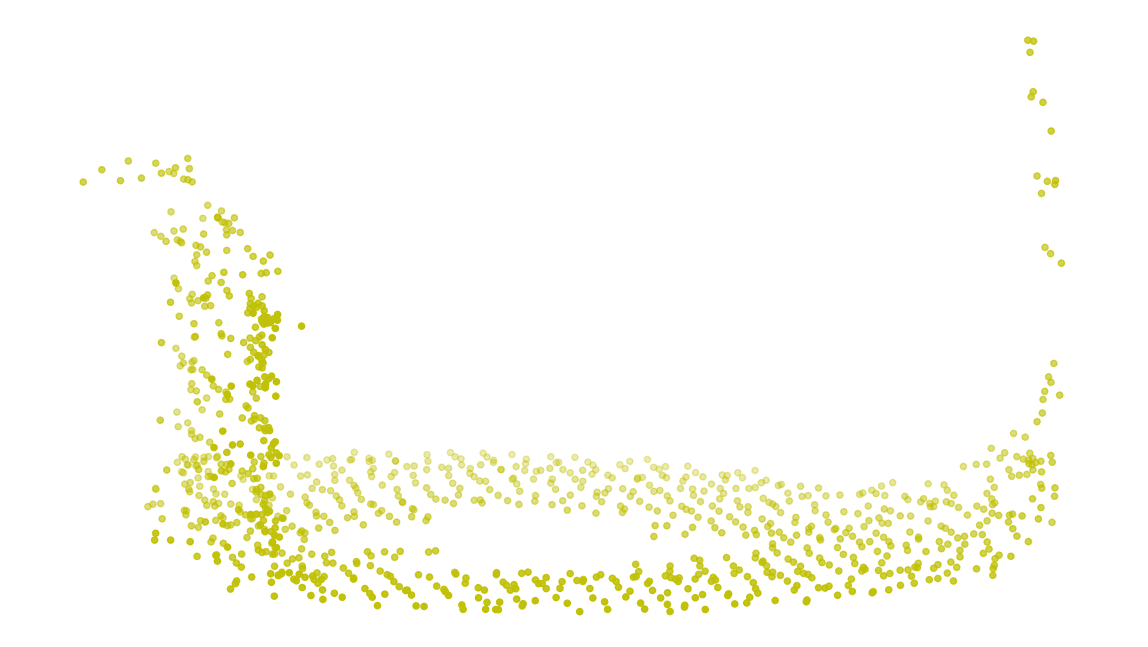}
    \caption{}  \label{subfig:unchanged_3_mjolk}
    \end{subfigure}
    \begin{subfigure}{0.28\textwidth}
    \includegraphics[width=\textwidth]{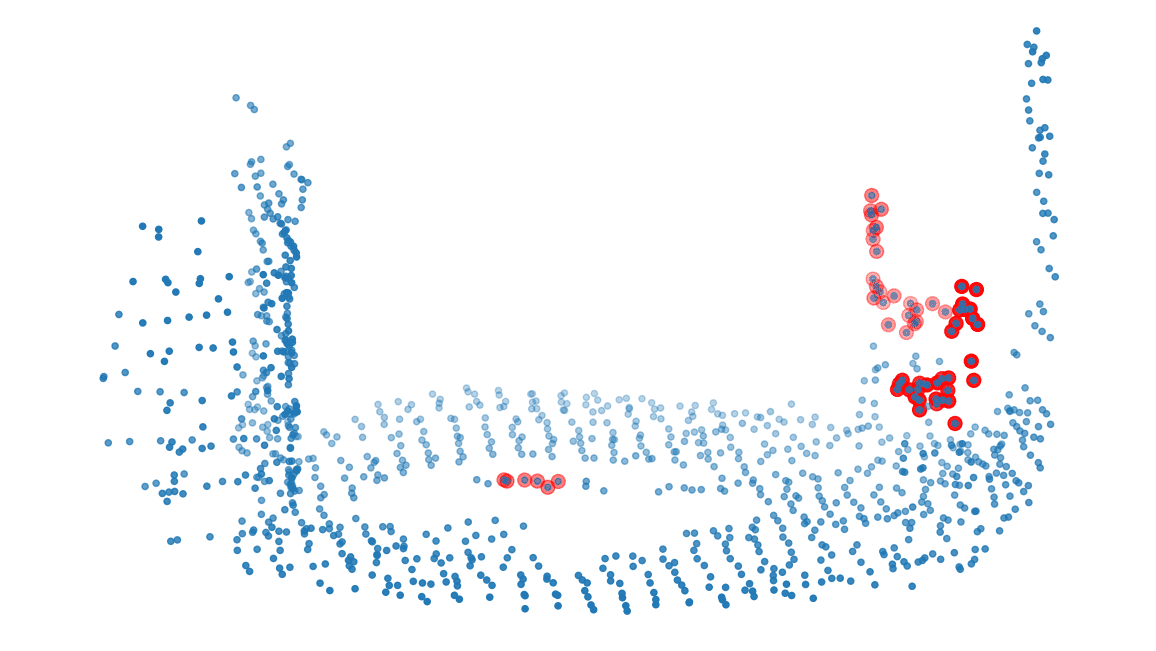}
     \caption{} \label{subfig:changed_3_mjolk}
    \end{subfigure}
    \begin{subfigure}{0.21\textwidth}
    \includegraphics[width=\textwidth]{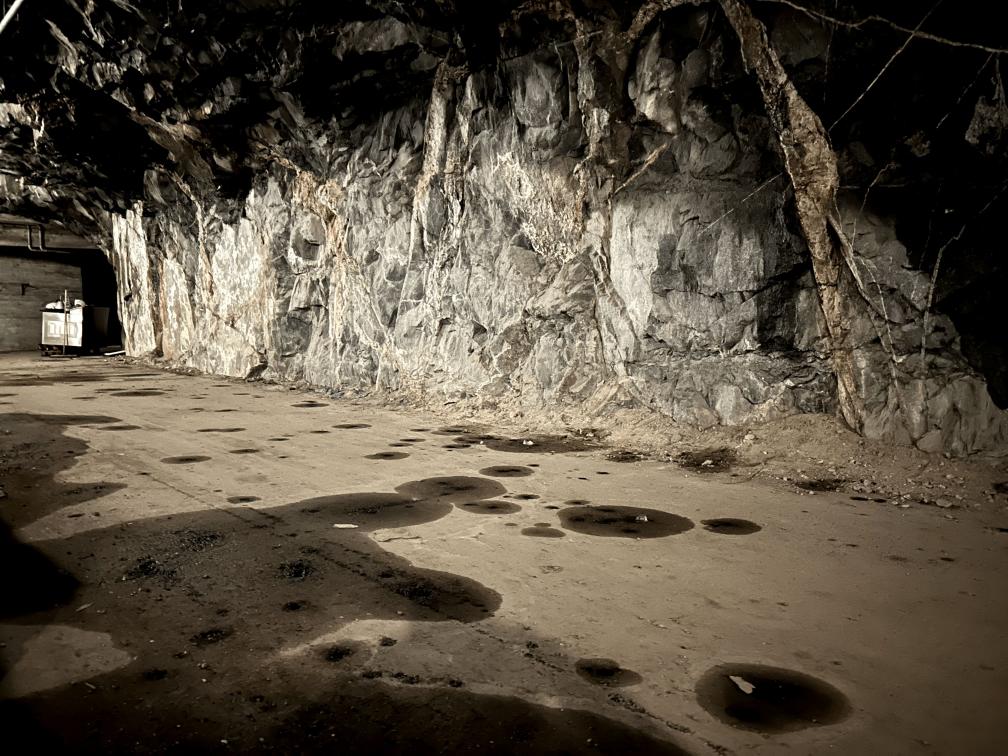} 
    \caption{}
    \end{subfigure}
    \hfill
    \begin{subfigure}{0.21\textwidth}
    \includegraphics[width=\textwidth]{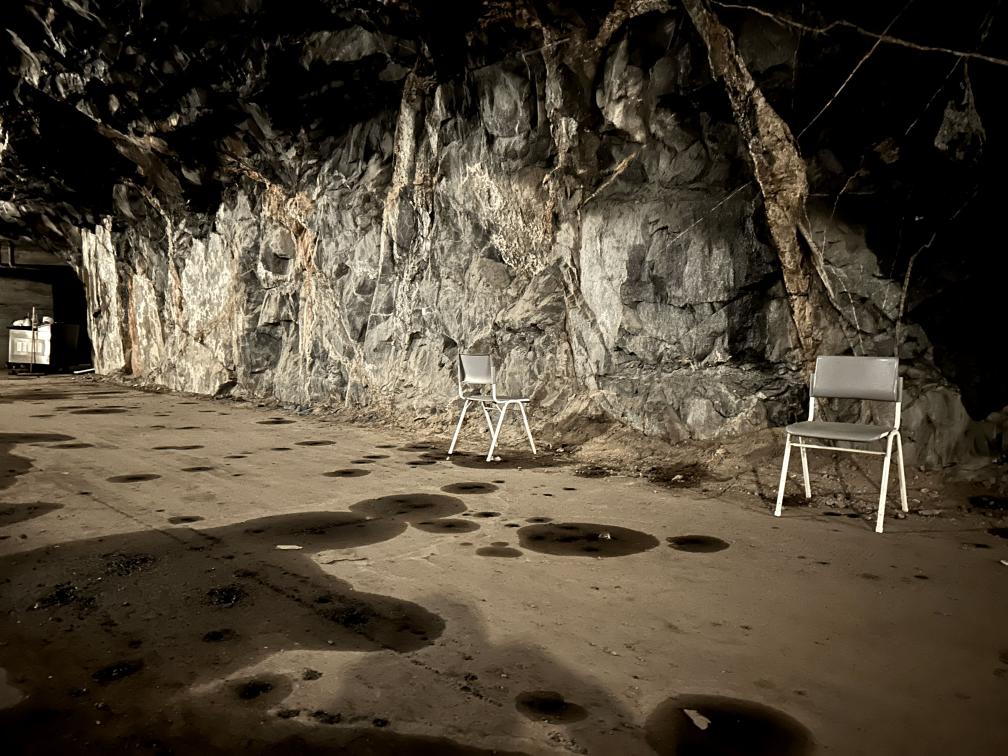} 
    \caption{}
    \end{subfigure}
    \caption{Visual comparison of the detected changed areas and their added objects. The first two columns demonstrate the point cloud representations, with and without the obstructions and the last two columns are the corresponding pictures from the testing environment.}
    \label{fig:changes}
\end{figure*}
%%%%%%%%%%%%%%%%%%%%%%%%%%%%%%%%%%%%%%%%%%%%%%%%%%%%%%%%%%%%%%%%%%%%%%%%%%%%%%%%
\subsection{Dataset and platform}
The framework was evaluated in two different subterranean environments, designed to test the method's ability to detect changes and extract objects from 3D point cloud scans. In both scenarios, the robot explores the area in two different time instances, starting roughly at the same position. The first environment, depicted on Fig.~\ref{fig:maps_mjolkberget}, is an underground tunnel network~\cite{kyuroson2023multimodal}, where at first, it does not contain any obstacles. Later, we manually place or move obstructions, as seen on Fig.~\ref{fig:changes}. The objects include boxes, chairs and a two wheel trolley. The robotic platform used is a mobile, wheeled, ground vehicle equipped with an OS1-32 LiDAR, featuring 32 laser channels with a $360^o \times 45^o$ horizontal and vertical FOV. The robot is carrying its own computational unit and all the processes are performed on-board the Intel NUC computer, with an Intel i7 CPU and 16~GB of RAM. The second environment, depicted on Fig.~\ref{fig:maps_epiroc}, is a 10 meter wide test mining site where we were able to test the proposed framework in real-world industrial conditions. At first the tunnel is obstacle free, while later on, obstructions to the sides of the wall were placed along with muck piles that were moved by a wheel loader. The robotic platform was a custom-made quadrotor, equipped with the same sensors and on-board processing unit as the aforementioned robot. 
% All the integrated algorithms are implemented using the ROS Noetic on Ubuntu 20.04.
% \vspace{-1mm}
%%%%%%%%%%%%%%%%%%%%%%%%%%%%%%%%%%%%%%%%%%%%%%%%%%%%%%%%%%%%%%%%%%%%%%%%%%%%%%%%
\subsection{Results}
%%%%%%%%%%%%%%%%%%%%%%%%%%%%%%%%%%%%%%%%%%%%%%%%%%%%%%%%%%%%%%%%%%%%%%%%%%%%%%%%
\begin{figure*}[t!]
  \begin{subfigure}{1.\columnwidth}
    \includegraphics[width=\columnwidth]{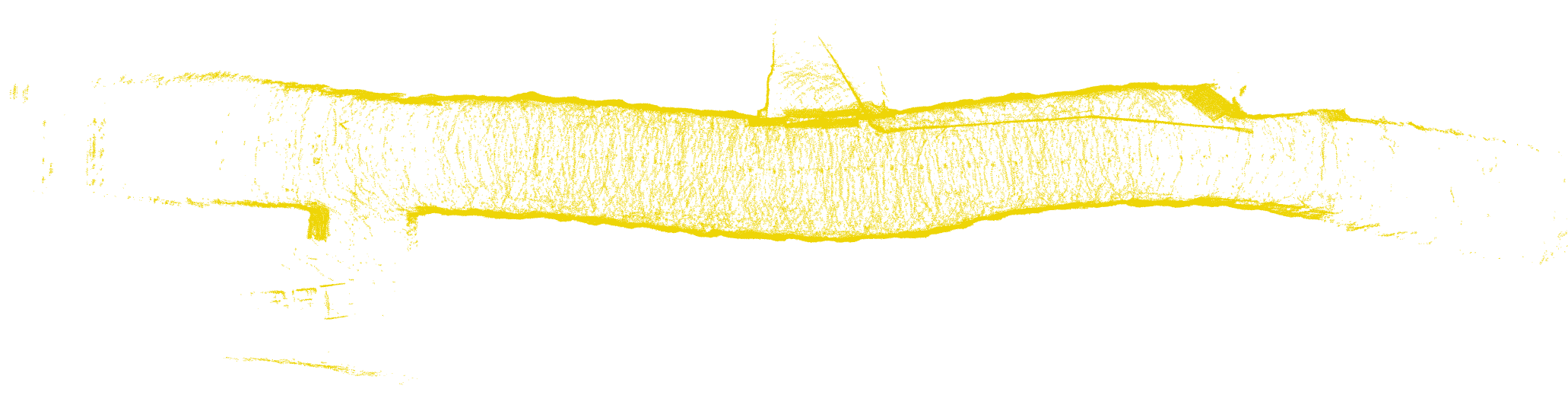}
    \caption{Map $M_t$}
    \end{subfigure}
    \hfill
    \begin{subfigure}{1.\columnwidth} 
    \includegraphics[width=\columnwidth]{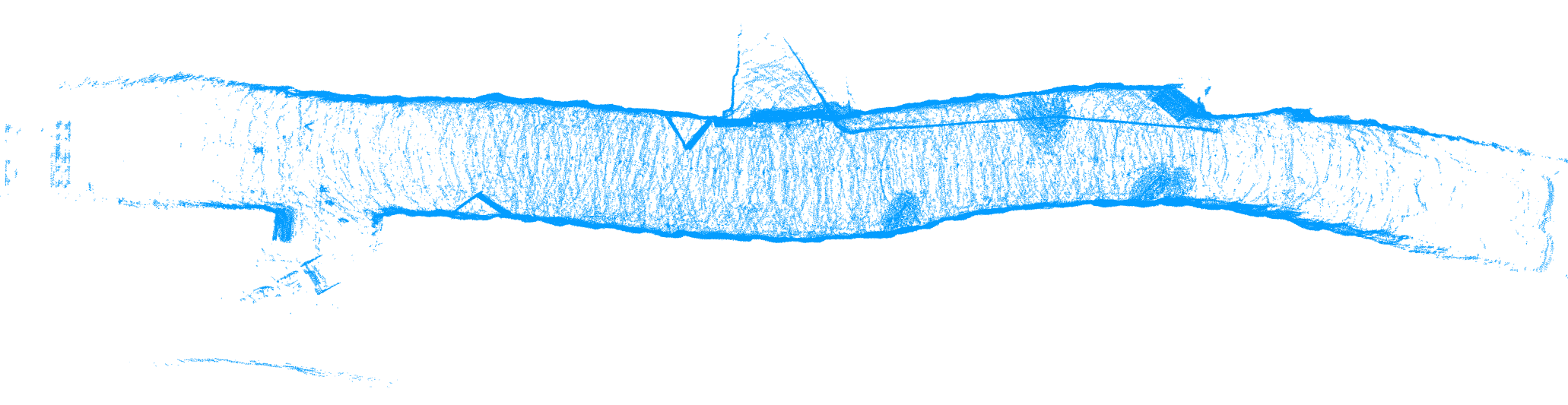}
    \caption{Map $M_{t+1}$}
    \end{subfigure}
    \vspace{5pt} \\
    \begin{subfigure}{1.\columnwidth} 
    \includegraphics[width=\columnwidth]{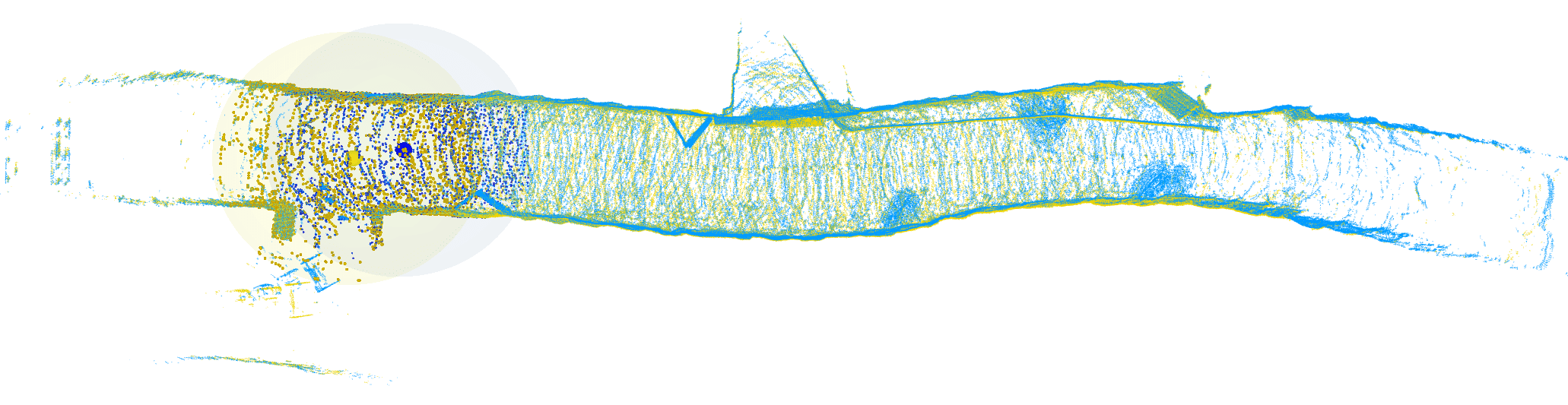}
    \caption{After alignment} \label{subfig:aligned_epiroc}
    \end{subfigure}
    \hfill
    \begin{subfigure}{1.\columnwidth}
    \includegraphics[width=\columnwidth]{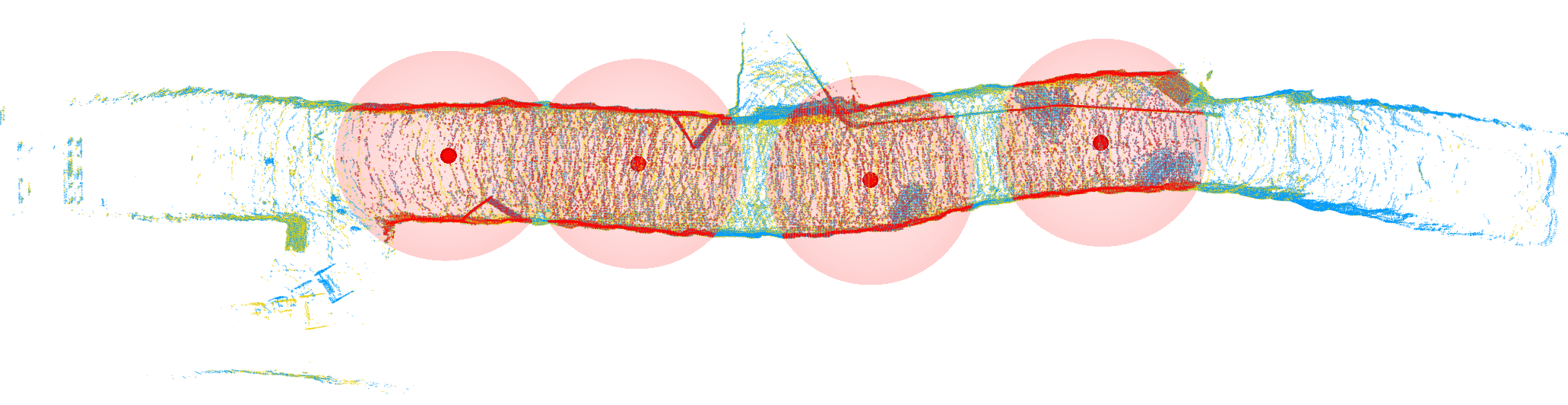}
    \caption{Detected changed areas} \label{subfig:changes_epiroc}
    \end{subfigure}
    \caption{Subfigure (a) depicts the map $M_t$ before the changes, consisting of a total of $1,263,631$ points, while subfigure (b) depicts the map $M_{t+1}$ after the changes, consisting of $1,439,038$ points. Subfigure (c) depicts the two maps after the alignment and subfigure (d) demonstrates the detected changed areas in red color.} \label{fig:maps_epiroc}
    \vspace{-4.mm}
\end{figure*}
%%%%%%%%%%%%%%%%%%%%%%%%%%%%%%%%%%%%%%%%%%%%%%%%%%%%%%%%%%%%%%%%%%%%%%%%%%%%%%%%
\begin{figure*}[b!]
\vspace{-5mm}
  \begin{subfigure}{0.24\textwidth}
    \includegraphics[width=\textwidth]{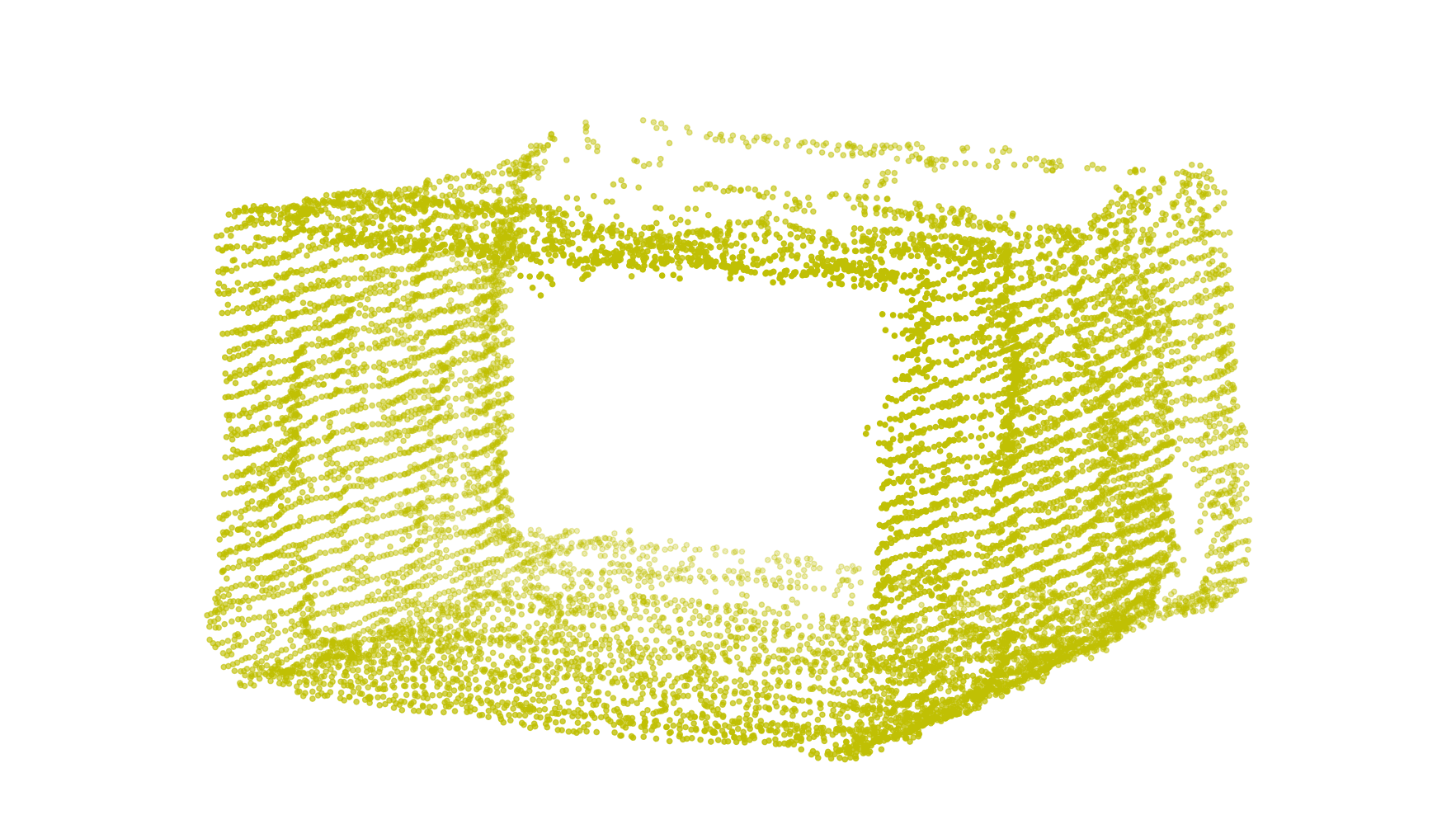}
    \caption{}
    \end{subfigure}
    \hfill
    \begin{subfigure}{0.24\textwidth}
    \includegraphics[width=\textwidth]{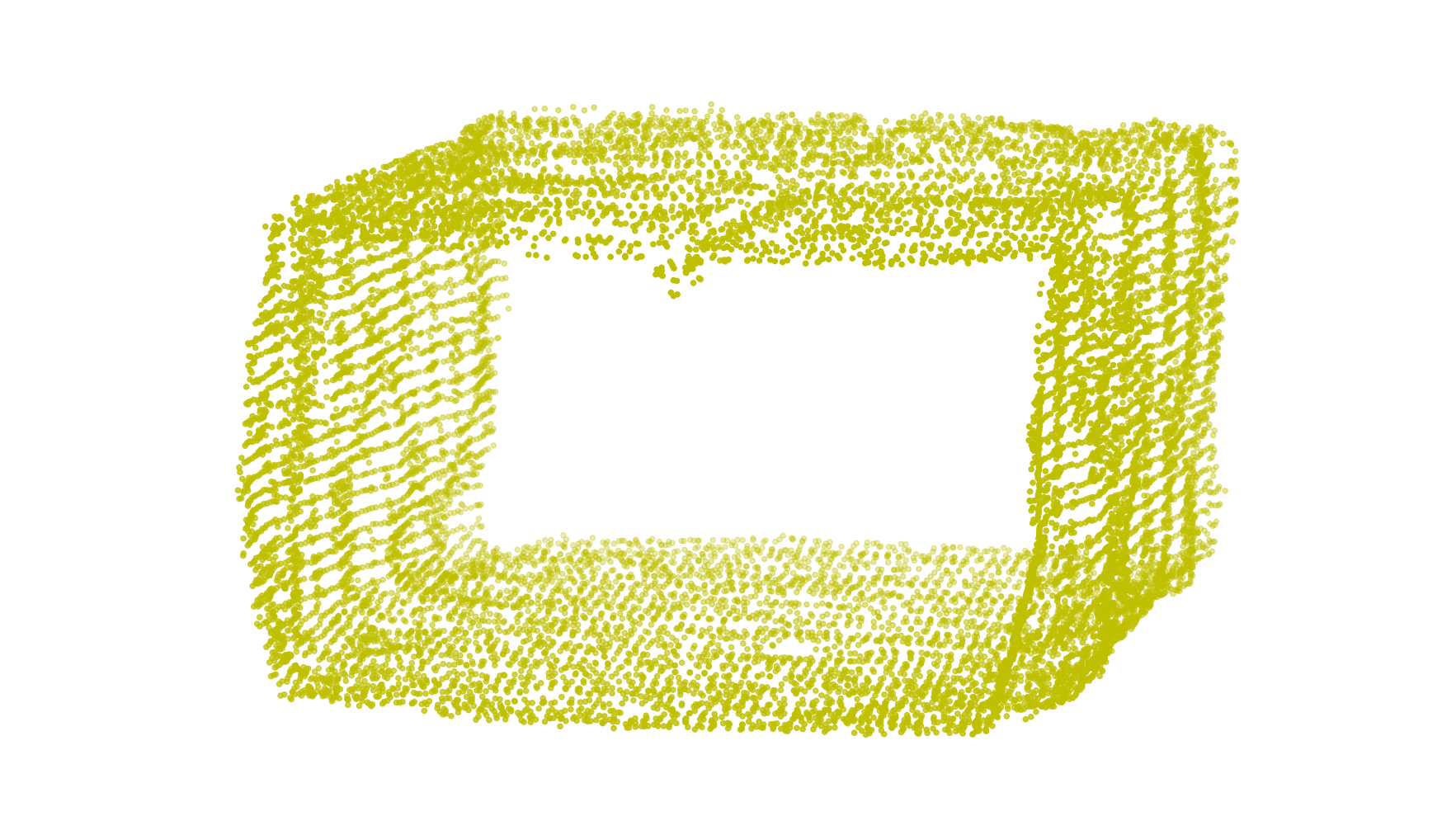}
     \caption{}
    \end{subfigure}
    \begin{subfigure}{0.24\textwidth}
    \includegraphics[width=\textwidth]{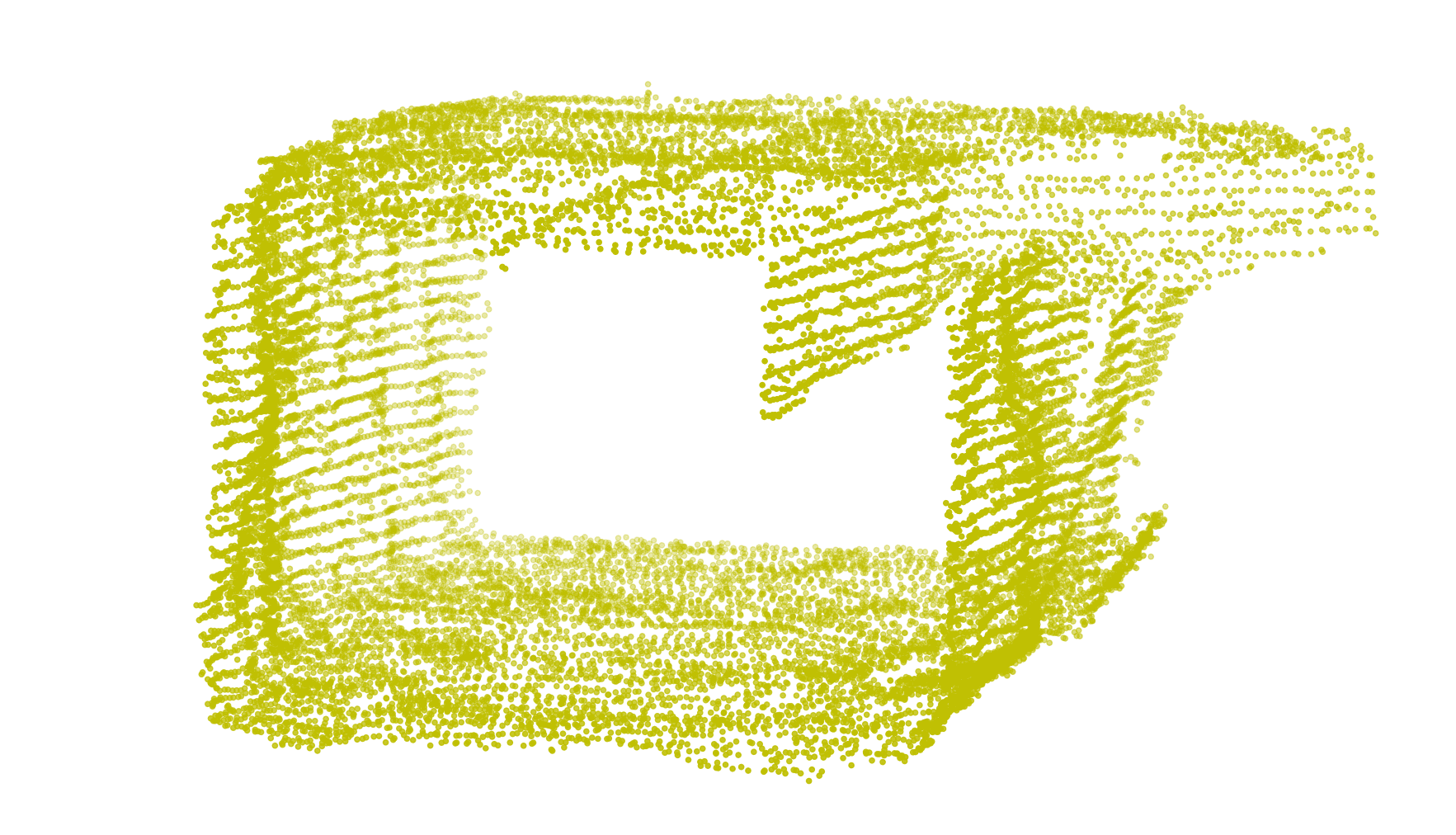} 
    \caption{}
    \end{subfigure}
    \hfill
    \begin{subfigure}{0.24\textwidth}
    \includegraphics[width=\textwidth]{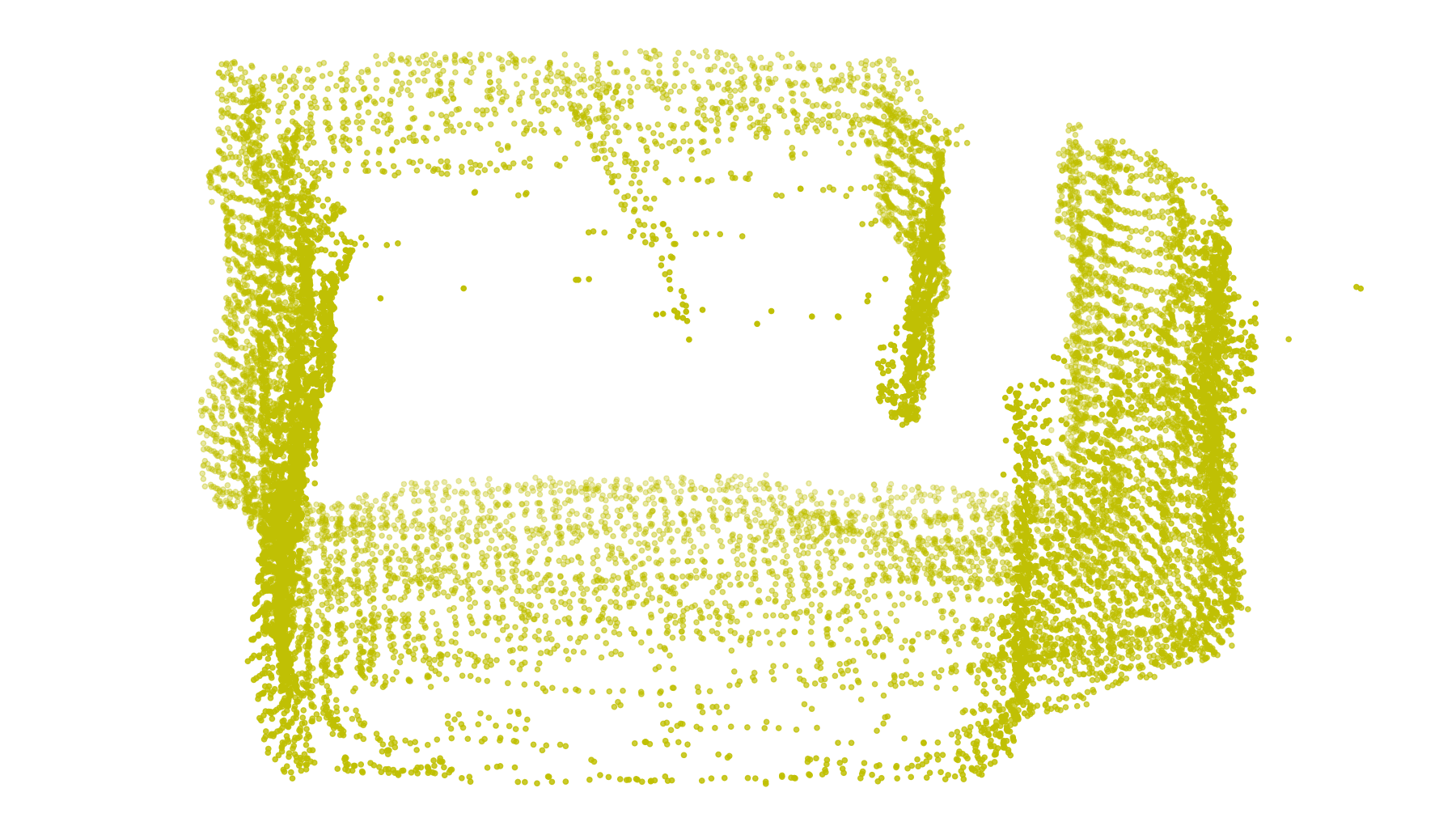} 
    \caption{}
    \end{subfigure}
    \begin{subfigure}{0.24\textwidth}
    \includegraphics[width=\textwidth]{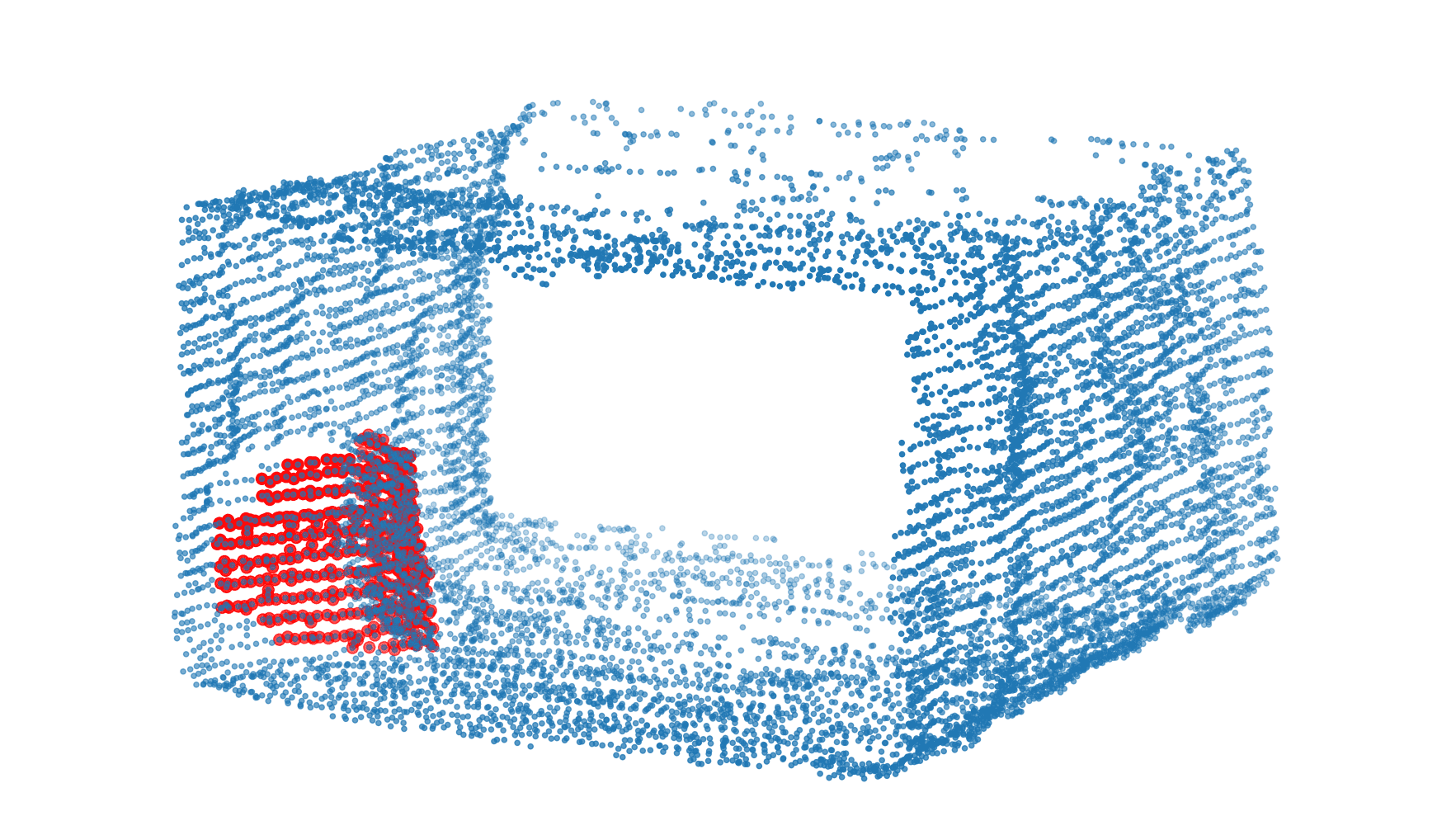}
    \caption{}
    \end{subfigure}
    \hfill
    \begin{subfigure}{0.24\textwidth}
    \includegraphics[width=\textwidth]{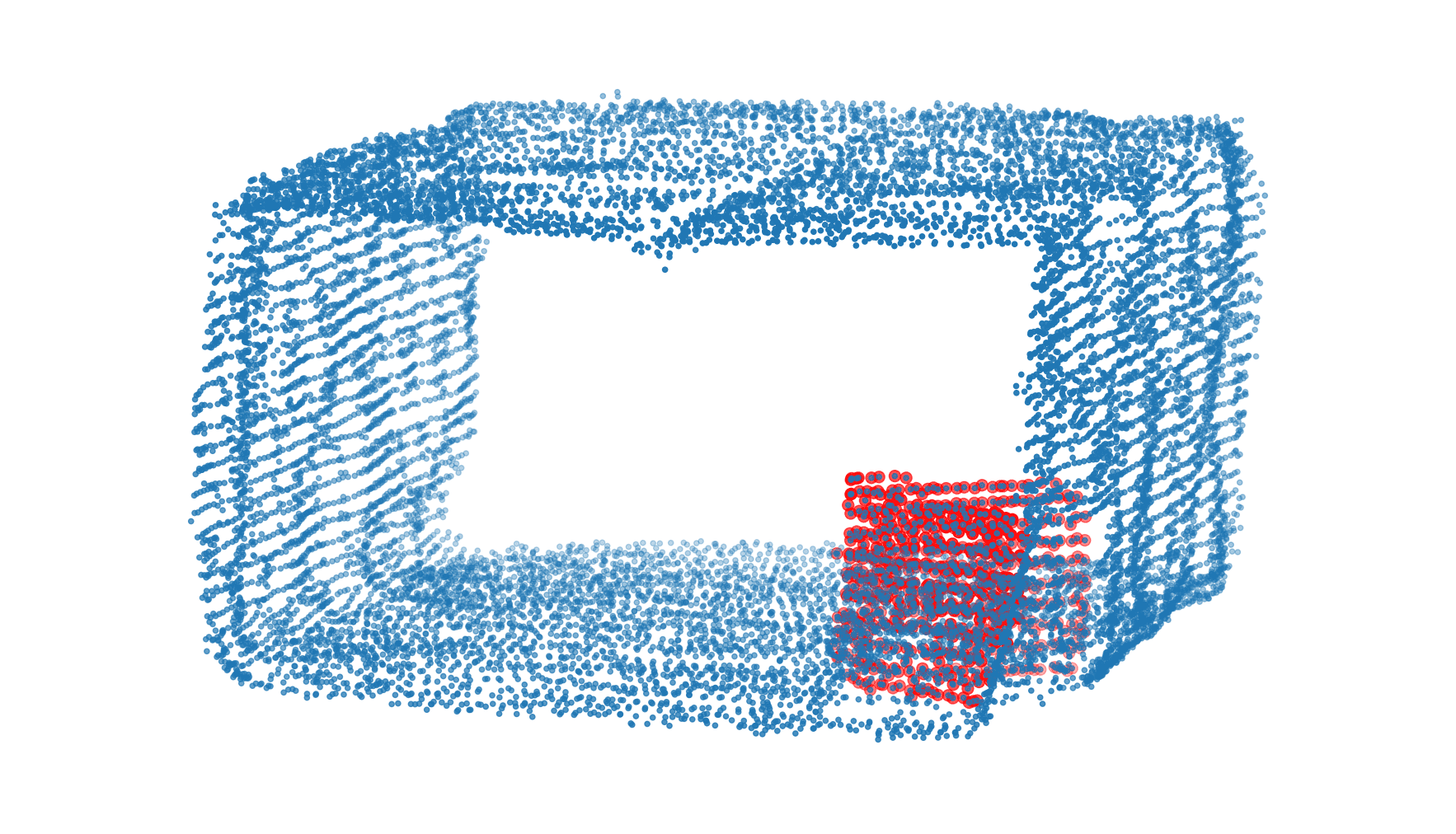}
     \caption{}
    \end{subfigure}
    \begin{subfigure}{0.25\textwidth}
    \includegraphics[width=\textwidth]{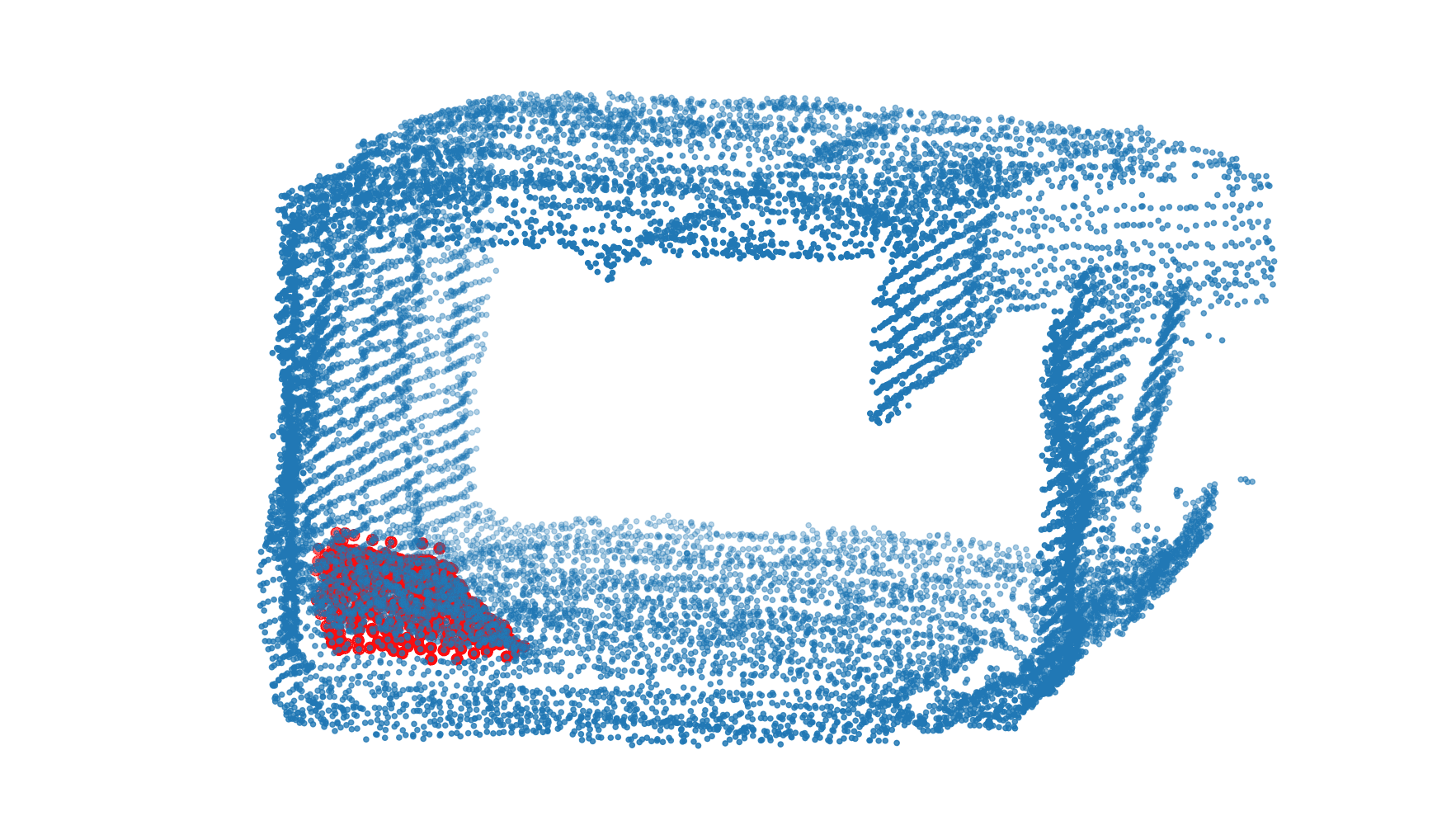}
    \caption{}
    \end{subfigure}
    \hfill
    \begin{subfigure}{0.24\textwidth}
    \includegraphics[width=\textwidth]{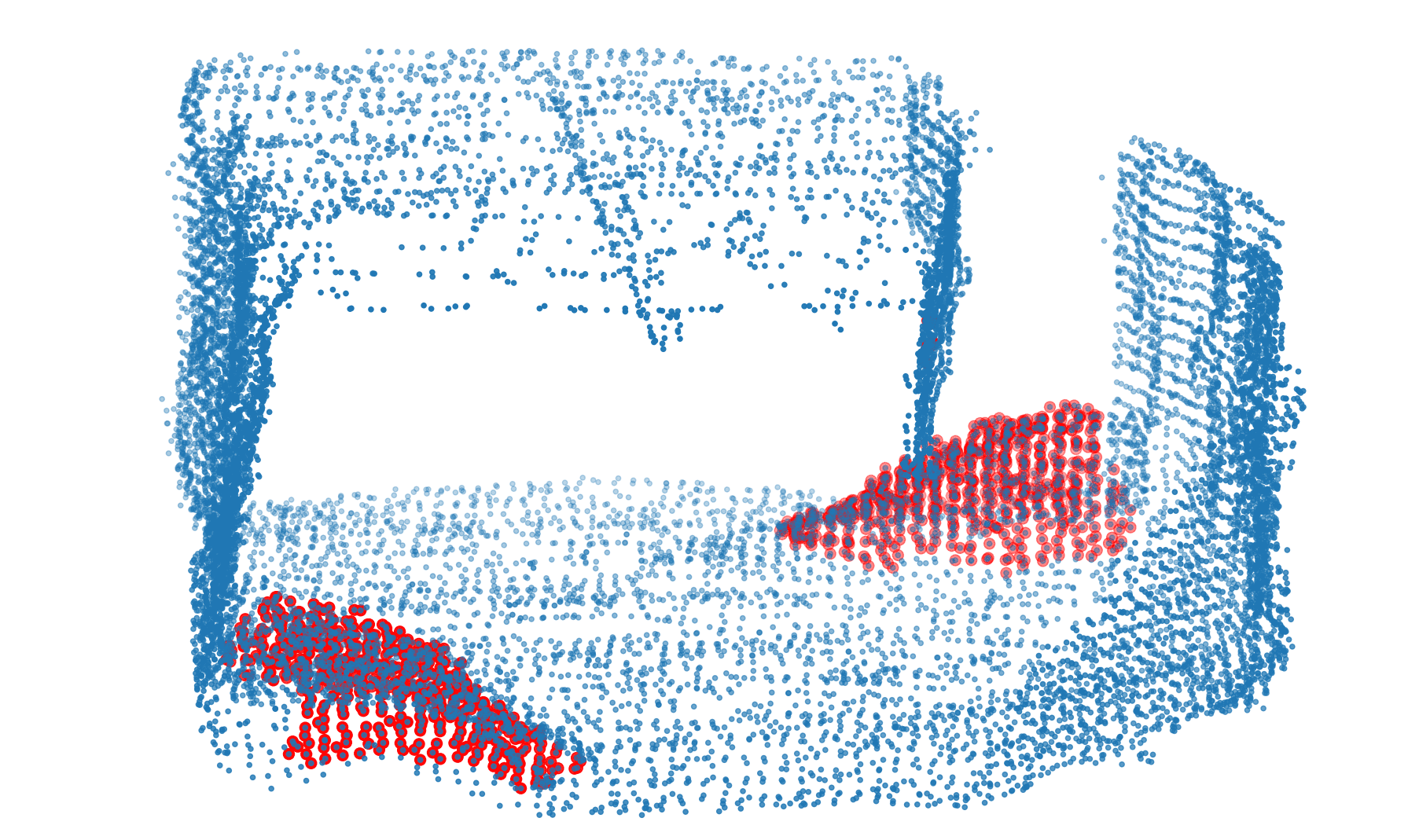}
     \caption{}
    \end{subfigure}
    \caption{Visual comparison of the detected changed areas and the extracted obstacles, represented with red points. The first row demonstrates the point cloud representations, before the addition of the obstructions and the second row demonstrates the same area, after the addition of the railings and the muckpiles.}
    \label{fig:changes_epiroc}
\end{figure*}
%%%%%%%%%%%%%%%%%%%%%%%%%%%%%%%%%%%%%%%%%%%%%%%%%%%%%%%%%%%%%%%%%%%%%%%%%%%%%%%%
The results of these experiments on change detection and object extraction are presented in the following figures. Starting from Fig.~\ref{fig:maps_mjolkberget} and the first environment, we can see the two maps $M_t$ and $M_{t+1}$ from two different time instances. While the first map is obstacle free, in the second one, two objects have been placed in various positions and another three have been moved. The first step of the proposed algorithm is to align the two maps into a single coordinate frame, $\mathcal{W}^G$, as depicted on Fig.~\ref{subfig:aligned_mjolk}. The processing time of this operation is just $51$ ms, displayed in Table~\ref{table:time}. Next, in the subfigure~\ref{subfig:changes_mjolk}, the three changed areas are detected and highlighted in red, indicating a good level of accuracy. By querying the two vector sets $Q_t$ and $Q_{t+1}$, searching for the maximum distance, the algorithm is able to extract all three areas even though the objects have a small volume compared to the total volume of a point cloud scan. The volume comparison can be seen in Table~\ref{table:volume}. After completing this step, the spheres $S_{k_i}$ and $S_{k_j}$ are sampled with a radius of $r=4.5$ meters, in order to move to the final stage of the object extraction. The process of converting the sphere $S_{k_i}$ to the voxel grid $V_{k_i}$ and performing a point-to-voxel comparison is efficient enough to keep the processing time under $11$ ms even for a sphere with the size of approximately $15,000$ points. The size of the voxel grid is set to $0.65$ meters throughout the whole experiment. The objects extracted from the detected changed areas are presented in Fig. \ref{fig:changes}, along with the corresponding pictures of the environment, with and without the objects. The proposed framework demonstrates a robust object extraction performance, being able to extract objects with as small of a volume as $0.30$ m$^3$, in a sphere with a volume of $206$ m$^3$.
As mentioned in section~\ref{sec:methodology}, the last step involves a statistical outlier removal filter. A limitation of the framework is evident and can be seen in subfigure~\ref{subfig:changed_3_mjolk}, where the filter is unable to remove all irrelevant points. If we apply a more aggressive filter, we risk of losing points within the object of interest, since in this experiment the objects are represented by very sparse point clouds, testing the limits of the proposed framework. 
Continuing from the previous experiment, the real mining site presented in Fig.~\ref{fig:maps_epiroc}, offers much wider tunnels and more realistic obstructions, like railings and muck piles. Furthermore, the 20-fold increase in map size compared to the previous experiment, approximately $60,000$ points compared to $1,400,000$ points, offers a good comparison for scalability purposes. Similar to the previous experiment, the first map $M_t$ is obstacle free while the second map, $M_{t+1}$ contains five obstacles in total, two railings and three muck piles. The first step of aligning the two maps into a single coordinate frame, $\mathcal{W}^G$, as depicted on Fig.~\ref{subfig:aligned_epiroc}, remains efficient with a processing time of $184$ ms, increasing only by a factor of 4. Next, in the subfigure~\ref{subfig:changes_epiroc}, the changed areas are detected and highlighted in red, demonstrating the algorithm's ability to handle a larger environment with the same level of accuracy. By querying the two vector sets $Q_t$ and $Q_{t+1}$, the algorithm can still extract all changed areas despite the increased volume of the point cloud map, as seen in Table~\ref{table:volume}. The spheres $S_{k_i}$ and $S_{k_j}$ are now sampled with a radius of $r=10$ meters in order to accommodate for the wider area, and the process of converting the sphere $S_{k_i}$ to the voxel grid $V_{k_i}$ and detecting the areas remains efficient, with processing times around $100$ ms. The objects extracted from the detected changed areas are presented in Fig. \ref{fig:changes_epiroc}. The proposed framework continues to demonstrate robust object extraction performance, handling the extraction of both railings and the muck piles, indicating future possibilities for usage in industrial sites similar to the testing environment. As previously mentioned, the last step involves a statistical outlier removal filter. In this case, the filter works effectively in removing all irrelevant points since the objects have a much greater volume as well as a higher point density, unlike in the previous experiment where a limitation of the framework was evident. 

%%%%%%%%%%%%%%%%%%%%%%%%%%%%%%%%%%%%%%%%%%%%%%%%%%%%%%%%%%%%%%%%%%%%%%%%%%%%%%%%
\begin{table}[h!]
\centering
\caption{The experimental results for the computational time of the map merging process ($t_{merge}$), the change detection ($t_{CD}$) and the object extraction ($t_{OE}$) processes as well as the total time ($t_{total}$)} \label{table:time}
\resizebox{1.\linewidth}{!}{%
\begin{tabular}{lcccc} 
\toprule
                 & $t_{merge}$ (s)         & $t_{CD}$ (s) & t$_{OE}$ (s) & $t_{total}$ (s)  \\ 
\toprule
Fig \ref{fig:changes}. (a) – (b) & \multirow{3}{*}{0.0510} & 0.0162       & 0.0005       & 0.0678           \\
Fig \ref{fig:changes}. (e) – (f) &                         & 0.0190       & 0.0004       & 0.0705           \\
Fig \ref{fig:changes}. (i) – (j) &                         & 0.0173       & 0.0004       & 0.0687           \\
Fig \ref{fig:changes_epiroc}. (a) – (b) & \multirow{4}{*}{0.1840} & 0.1085       & 0.0057       & 0.1142           \\
Fig \ref{fig:changes_epiroc}. (c) – (d) &                         & 0.1149       & 0.0078       & 0.1227           \\
Fig \ref{fig:changes_epiroc}. (e) – (f) &                         & 0.1112       & 0.0072       & 0.1185           \\
Fig \ref{fig:changes_epiroc}. (g) – (h) &                         & 0.1087       & 0.0110       & 0.1197           \\
\toprule
\end{tabular}
}
\vspace{-5mm}
\end{table}
%%%%%%%%%%%%%%%%%%%%%%%%%%%%%%%%%%%%%%%%%%%%%%%%%%%%%%%%%%%%%%%%%%%%%%%%%%%%%%%%
%%%%%%%%%%%%%%%%%%%%%%%%%%%%%%%%%%%%%%%%%%%%%%%%%%%%%%%%%%%%%%%%%%%%%%%%%%%%%%%%
\begin{table}[h!]
\centering
\caption{Comparison of the approximate volume of the sampled spherical regions ($V_{sphere}$) and the objects extracted ($V_{OE}$), along with their corresponding number of points ($S_{points}$ and $OE_{points}$).} \label{table:volume}
\resizebox{1.\linewidth}{!}{%
\begin{tabular}{lcccc} 
\toprule
                 & $V_{sphere}\;(m^3)$ & $V_{OE}\;(m^3$) & ~$S_{points}$ & $OE_{points}$  \\ 
\toprule
Fig \ref{fig:changes}. (a) – (b) & 206                & 0.30            & 2,470         & 47             \\
Fig \ref{fig:changes}. (e) – (f) & 223                & 0.51            & 1,590         & 71             \\
Fig \ref{fig:changes}. (i) – (j) & 212                & 0.38            & 648           & 25             \\
Fig \ref{fig:changes_epiroc}. (a) – (b) & 1,452              & 22              & 11,391        & 557            \\
Fig \ref{fig:changes_epiroc}. (c) – (d) & 1,423              & 25              & 15,749        & 607            \\
Fig \ref{fig:changes_epiroc}. (e) – (f) & 1,564              & 12              & 16,186        & 470            \\
Fig \ref{fig:changes_epiroc}. (g) – (h) & 1,589              & 38              & 15,491        & 673            \\
\toprule
\end{tabular}
}
\vspace{-5mm}
\end{table}
%%%%%%%%%%%%%%%%%%%%%%%%%%%%%%%%%%%%%%%%%%%%%%%%%%%%%%%%%%%%%%%%%%%%%%%%%%%%%%%%

%%%%%%%%%%%%%%%%%%%%%%%%%%%%%%%%%%%%%%%%%%%%%%%%%%%%%%%%%%%%%%%%%%%%%%%%%%%%%%%%
\section{Conclusions} \label{sec:conclusions}
%%%%%%%%%%%%%%%%%%%%%%%%%%%%%%%%%%%%%%%%%%%%%%%%%%%%%%%%%%%%%%%%%%%%%%%%%%%%%%%%

In conclusion, the novel proposed approach for change detection and irregular object extraction in sparse bi-temporal  3D point clouds using deep learned place recognition descriptors and voxel-to-point comparison is a promising solution for monitoring dynamic structured environments. The approach successfully detected changes in real-world field experiments and demonstrated its effectiveness in detecting different types of changes, such as object or muck-pile addition and displacement. The approach has important implications and provides new insights for various applications, including safety and security monitoring in construction sites, mapping and exploration. The study suggests potential future research directions in this field, such as extending the proposed approach to larger datasets or exploring the potential of incorporating other deep learning techniques aiming towards real-time solutions. Overall, this research offers a valuable contribution to the field of change detection and abnormal object extraction in sparse 3D point clouds.

%%%%%%%%%%%%%%%%%%%%%%%%%%%%%%%%%%%%%%%%%%%%%%%%%%%%%%%%%%%%%%%%%%%%%%%%%%%%%%%%

% \addtolength{\textheight}{-15cm}   % This command serves to balance the column lengths

\bibliographystyle{./IEEEtranBST/IEEEtran}
\bibliography{./IEEEtranBST/IEEEabrv,root}

\end{document}